\documentclass{article}

\usepackage{arxiv}

\usepackage[super,sort,comma]{natbib}
\usepackage{amsmath,amsfonts,amssymb}
\usepackage{graphicx}
\usepackage[colorlinks=true, allcolors=blue]{hyperref}
\usepackage{multirow}
\usepackage{booktabs}
\usepackage{subcaption}
\usepackage{epsfig}
\usepackage{bbm}
\usepackage{fancyhdr}	
\usepackage{dsfont}
\usepackage{caption}
\title{Improving PET/CT-Based Whole-Body Lesion Segmentation Using Prediction Uncertainty-Augmented Models}

\author{ Bashirul Azam Biswas \\
Department of Biomedical Data Science\\
Geisel School of Medicine at Dartmouth\\
Hanover, NH 03755, USA\\
	\texttt{Bashirul.Azam.Biswas@dartmouth.edu} \\
	\And
	Biratal Raj Wagle \\
Department of Biomedical Data Science\\
Geisel School of Medicine at Dartmouth\\
Hanover, NH 03755, USA\\
	\texttt{Biratal.Raj.Wagle@dartmouth.edu} \\
    \And
	Zhihan Yang \\
Department of Biomedical Data Science\\
Geisel School of Medicine at Dartmouth\\
Hanover, NH 03755, USA\\
	\texttt{Zhihan.Yang.GR@dartmouth.edu} \\
    \And
	Marc A. Seltzer \\
Radiology\\
Dartmouth Hitchcock Medical Center\\
Lebanon, NH 03766 , USA\\
	\texttt{marc.a.seltzer@hitchcock.org} \\
    \And
	Matthew E. Maeder \\
Radiology\\
Dartmouth Hitchcock Medical Center\\
Lebanon, NH 03766 , USA\\
	\texttt{Matthew.E.Maeder@dartmouth.edu}\\ 
    \And
	James B. Yu \\
Radiation Oncology\\
Dartmouth Hitchcock Medical Center\\
Lebanon, NH 03766 , USA\\
	\texttt{James.B.Yu@dartmouth.edu}\\ 
    \And
	Indrani Bhattacharya \\
Department of Biomedical Data Science\\
Geisel School of Medicine at Dartmouth\\
Hanover, NH 03755, USA\\
	\texttt{Indrani.Bhattacharya@dartmouth.edu}\\
}

\renewcommand{\shorttitle}{Improving PET/CT-based WBLS Using Uncertainy-Augmented Models}

\hypersetup{
pdftitle={Improving PET-CT-based WBLS with Uncertainty},
pdfsubject={cs,eecs},
pdfauthor={Bashirul},
pdfkeywords={Whole-body lesion segmentation, Uncertainty quantification, Uncertainty-aware modelling},
}

\begin{document}
\maketitle

% keywords can be removed

\begin{abstract}
\noindent {\bf Background:} Accurate lesion segmentation from whole-body Positron Emission Tomography (PET)/Computed Tomography (CT) scans is essential for cancer staging and treatment planning. PET provides functional metabolic information with different radiotracers (e.g., $[^{18}F]$fluorodeoxyglucose (FDG),  prostate specific membrane antigen (PSMA)), while CT is primarily used for attenuation correction and offers anatomical localization. Lesion delineation from PET/CT imaging is clinically challenging due to subtle imaging features, confounders, and inter-reader variability. The rapid increase in PET/CT imaging without a corresponding increase in subspecialty-trained radiologists to interpret them poses significant burden on radiologist workloads and can inhibit timely treatment planning in high-risk cancer patients. 
\\ 
{\bf Purpose:} There is a clinical need for accurate and generalizable automated whole-body PET/CT lesion segmentation methods that can assist radiologists in pan-cancer, multi-tracer images. Existing deep learning approaches suffer from training-related stochasticity, inconsistent predictions, missed lesions in high tumor-burden cases, and lack prediction uncertainty quantification, limiting their clinical reliability and adoption. \\
{\bf Methods:} Using the nnU-Net model as baseline, we propose an uncertainty-aware framework for PET/CT whole-body lesion segmentation that integrates (1) Bayesian ensembling to mitigate training stochasticity, (2) voxel-wise uncertainty quantification and decomposition in epistemic (model-specific) and aleatoric (data-specific) components, and (3) epistemic uncertainty-augmented training to improve lesion detection. We analyze the relationship between epistemic uncertainty and different types of model misclassification, and assess the ability of uncertainty estimates to identify misclassified voxels. We also evaluate the generalizability of our method on unseen datasets. Two public datasets: AutoPET-III (1,611 scans) and Deep-PSMA (200 scans), both comprising FDG and PSMA studies across multiple cancer types, are used for our training and evaluation. \\
{\bf Results:}
Bayesian ensembling improves robustness and performance over deterministic nnU-Net models on the unseen AutoPET-III test set (Dice: 60.1 $\rightarrow$ 62.0; False Positive Volume (FPVol): 2.9 $\rightarrow$ 2.0; False Negative Volume (FNVol): 13.9 $\rightarrow$ 13.6). Uncertainty maps highlight regions of model disagreement and correlate with misclassifications, with higher uncertainty observed for missclassified predictions, particularly false positives. Uncertainty-augmented training enhances lesion recovery (FNVol: 13.6 $\rightarrow$ 9.3) at the cost of increased FPVol (2.0 $\rightarrow$ 4.8), reflecting a precision–recall trade-off. A case-adaptive routing strategy improves Dice performance with statistical significance by selecting between base and augmented models (baseline Bayesian $\rightarrow$ augmented Bayesian $\rightarrow$ case-adaptive routing: 62.0 $\rightarrow$ 61.7 $\rightarrow$ 63.1 ). On the out-of-distribution Deep-PSMA dataset, Bayesian ensembling improves Dice (baseline $\rightarrow$ ours: 62.8 $\rightarrow$ 64.7), and the uncertainty-augmented model achieves the highest Dice (65.3) with a similar trade-off pattern. \\
{\bf Conclusions:} Bayesian ensembling enhances robustness and consistency in PET/CT whole-body lesion segmentation, while uncertainty quantification provides meaningful indicators of prediction reliability. Incorporating uncertainty into model training improves lesion detection by reducing false negatives. The proposed framework demonstrates strong generalization across both in-distribution and out-of-distribution datasets. To our knowledge, this is the first study to systematically investigate uncertainty quantification in multi-tracer, pan-cancer PET/CT segmentation and to integrate Bayesian ensembling with uncertainty-aware modeling for this task. \\

\end{abstract}


%\newpage     %may or may not be needed

%\newpage

\setlength{\baselineskip}{0.7cm}      %double spacing		

\pagenumbering{arabic}
\setcounter{page}{1}
\pagestyle{fancy}

\section{Introduction}\label{sec:intro}
Whole-body Positron Emission Tomography (PET)/Computed Tomography (CT) imaging is a valuable multimodal tool for cancer staging, treatment planning, and response assessment \cite{trotter2023positron}. PET visualizes physiological and molecular processes by detecting uptake of different radiotracers, such as $[^{68}\mathrm{Ga}]$-PSMA-11 and $[^{18}\mathrm{F}]$-labeled prostate specific membrane antigen (PSMA)-targeted radiotracers \cite{silver1997prostate}, $[^{18}F]$fluorodeoxyglucose ($[^{18}]$FDG) \cite{som1980fluorinated} and Sodium Fluorine-18 Fluoride ($[^{18}]$F-NaF) \cite{blau1962fluorine}. These PET radiotracers' uptake reflect underlying tissue metabolism or receptor activity, whereas CT provides complementary anatomical information. However, interpreting PET/CT imaging is challenging due to subtle multimodal imaging features, difficulty differentiating normal uptake from cancer, diversity of anatomical regions, different radiotracers, and the labor- and expertise-intensive task of lesion identification and annotation \cite{hofman2016we}. The growing volume of PET/CT scans, without a corresponding increase in nuclear medicine experts available for interpretation, creates a clinical need for standardized and generalizable computer-aided whole-body lesion segmentation systems. In addition, PET/CT scans are increasingly used for radiation treatment planning by radiation oncologists who may not have nuclear medicine fellowship training, often without explicit expert annotation of lesion extent or burden. As the number of radiotracers developed and used for cancer detection and staging continues to expand, there is a growing need for robust, reliable, and generalizable multi-tracer, pan-cancer PET/CT-based whole-body lesion segmentation methods that can provide diagnostic support across diverse diseases and imaging agents. 

Early computer-aided lesion segmentation methods for PET/CT employed diverse algorithms, including K-nearest neighbor classifiers \cite{yu2008coregistered}, Markov Random Fields (MRFs) \cite{yang2015multimodality}, subgraph-based energy minimization \cite{song2013optimal}, and hypergraph-based random walks \cite{bagci2013joint}. While these classical machine-learning approaches rely on hand-crafted feature extraction, deep learning (DL) methods later emerged, offering automatic feature learning along with improved and more generalizable segmentation performance \cite{huang2018fully}. Initial DL–based studies using PET/CT imaging primarily focused on specific anatomical regions, such as the head and neck \cite{huang2018fully, oreiller2022head}, lungs \cite{zhao2018tumor}, or esophagus \cite{jin2019accurate}. Huang \textit{et al.} \cite{huang2018fully} proposed a deep convolution network with downsampling blocks of feature representation and upsampling blocks of reconstruction for Gross Tumor Volume (GTV) segmentation in the head and neck region. Oreiller \textit{et al.} \cite{oreiller2022head} introduced a Head and Neck (H\&N) GTV detection challenge with FDG PET-CT imaging.  Zhao \textit{et al.} \cite{zhao2018tumor} adopted a feature fusion strategy from PET and CT segmentation network to render the final segmentation map for lung cancer. Jin \textit {et al.} \cite{jin2019accurate} integrated early and late fusion of 3D deep neural networks to perform esophageal GTV segmentation on PET/CT data. 

In addition to local/regional lesion segmentation, automated lesion segmentation from whole-body PET/CT imaging \cite{blanc2021fully, xu2018automated,jemaa2020tumor,he2023whole} has also been an active area of research. Blanc-Durand \textit{et al.} \cite{blanc2021fully} deployed a 3D U-Net architecture to detect and segment lymphoma lesions from whole-body FDG-PET/CT.  Jemaa \textit{et al.} \cite{jemaa2020tumor} developed a segmentation pipeline that integrates 2D segmentation, connected-component labeling, and a refinement stage leveraging 3D segmentation for enhanced accuracy with FDG PET/CT scans of whole-body. Gatidis \textit{et al.} \cite{gatidis2024results} introduced the AutoPET-II challenge and released a public dataset of FDG-PET/CT whole-body scans encompassing multiple tumour types to promote research in whole-body tumor segmentation. Utilizing this public dataset, later on, He \textit{et al.} \cite{he2023whole} deployed  Camouflaged Object Detection mechanisms and Liu \textit{et al.} \cite{liu2024lm} proposed hybrid Mamba-CNN architectures in lesion detection from whole-body scans. While FDG-PET is the most commonly used modality in PET/CT-based machine learning methods for whole-body lesion segmentation, other radiotracers such as $^{68}$Ga-Pentixafor \cite{xu2018automated} and PSMA \cite{xu2023automatic, li2024automated} have also been explored. Xu \textit{et al.}  \cite{xu2018automated} developed a W-Net by cascading two V-Nets \cite{milletari2016v}, with the first network trained on CT images and the second trained on combined $^{68}$Ga-Pentixafor PET and CT modalities to detect multiple Myeloma from whole-body scans. Li \textit{et al.} \cite{li2024automated} adopted 3D hybrid Transformer–CNN  and Xu \textit{et al.} \cite{xu2023automatic} deployed UNet-based architectures for whole-body PSMA-PET/CT lesion segmentation. All these PET/CT-based methods operate on local or whole-body anatomical scans and are designed to segment either single or multiple tumor types. 

One limitation of these prior studies is that they were limited to single-tracer datasets and lack generalization across multiple tracers. The recent AutoPet-III grand challenge  \cite{Ingrisch2024_autoPETIII} was organized to address this gap through the development of whole-body, multi-tracer (FDG and PSMA), pan-cancer (prostate, lung, melanoma, or lymphoma) segmentation methods. Most of the methods that were reported as part of this grand challenge  \cite{kalisch2024autopet,song2024dueu,rokuss2024fdg,chan2024autopet,wang2024dual}, used U-Net \cite{ronneberger2015u} or nnU-Net \cite{isensee2021nnu} variants. NnU-Net is a widely adopted medical image segmentation framework with strong performance across numerous tasks, including multi-tracer, pan-cancer PET/CT lesion segmentation. However, it still suffers from the following shortcomings:

\begin{enumerate} \itemsep -2pt
    \item \textbf{Sensitive to initialization and data-batching:} NnU-Net contains millions of parameters optimized through mini-batch training, making convergence sensitive to parameter initialization and batching strategies \cite{glorot2010understanding,keskar2016large}. As a result, models trained on the same dataset can produce different predictions for the same test case, posing a challenge for clinical deployment of whole-body lesion segmentation. Although Schott \textit{et al.} \cite{schott2025uncertainty} studied random-initialization ensembling for uni-tracer, uni-cancer PET/CT segmentation, they did not assess the impact of initialization and mini-batch variability on accuracy, precision, and recall in a multi-tracer, pan-cancer setting.

    \item \textbf{Absence of prediction uncertainty quantification in multi-tracer, pan-cancer model:} Deep learning segmentation models such as nnU-Net output voxel-wise lesion probabilities, which are often treated as confidence scores but do not reliably capture true model uncertainty \cite{kendall2017uncertainties,yu2024heterogeneity}. Uncertainty quantification provides a more principled estimate of prediction reliability by decomposing uncertainty into aleatoric and epistemic components \cite{lakshminarayanan2017simple,abdar2021review,kendall2017uncertainties}. Existing approaches include Bayesian methods such as Monte Carlo dropout, deep ensembles, and variational inference, as well as non-Bayesian methods such as Dempster–Shafer theory and test-time augmentation \cite{gal2016dropout,lakshminarayanan2017simple,zhao2022efficient,sensoy2018evidential,wang2019aleatoric}. Uncertainty quantification for whole-body PET/CT lesion segmentation remains limited. To the best of our knowledge, the recent study by Schott \textit{et al.} \cite{schott2025uncertainty} is the only study to investigate uncertainty estimation for nnU-Net on whole-body PET/CT and showed its utility for image degradation detection, False Positive (FP) identification, and False Negative (FN) recovery. However, their work is restricted to a single tracer and lesion type. Uncertainty quantification for multi-tracer, pan-cancer, whole-body PET/CT lesion segmentation remains underexplored, despite its importance for handling variability in lesion burden, cancer type, tracer distribution, scanner protocols, and multimodal PET/CT inputs. 
    
\item \textbf{Lack of uncertainty-aware modeling in multi-tracer, pan-cancer setting :} Our analysis of AutoPET-III shows that PSMA cases have more lesions than FDG cases (PSMA: 35.36 $\pm$ 51.54 vs. FDG: 12.42 $\pm$ 19.06) and higher, more variable lesion-wise intensities (PSMA: 7.56 $\pm$ 7.76 vs. FDG:4.95 $\pm$ 2.64). These differences are statistically significant by unpaired t-tests \cite{ruxton2006unequal} and Mann–Whitney U tests \cite{nachar2008mann}, with all p-values $<0.05$. We present tracer-specific distributions of lesion counts per case and mean lesion intensities for both FDG and PSMA in Figure \ref{figs:stat_tracer}. The AutoPET-III contains fewer PSMA samples ($597$) than FDG samples ($1097$), creating an approximately 1:2 imbalance. This can bias models toward FDG-like, lower-lesion-burden cases, promoting precision-oriented behavior and under-segmentation in PSMA, resulting in higher False Negative Volume (FNVol). Prior AutoPET-III methods similarly report lower PSMA performance and higher FNVol \cite{kovacs2024data,rokuss2024fdg,kalisch2024autopet}. Although Schott \textit{et al.} \cite{schott2025uncertainty} studied uncertainty for FP/FN analysis in whole-body PET/CT, their work is limited to single-tracer, single-cancer data and does not integrate uncertainty into nnU-Net for improving performance. Thus, uncertainty-aware modeling remains underexplored for heterogeneous multi-tracer, pan-cancer PET/CT lesion segmentation.

\begin{figure}[htbp]
	\centering
	\begin{subfigure}[b]{0.45\textwidth}
	\centering
	\includegraphics[width=\textwidth]{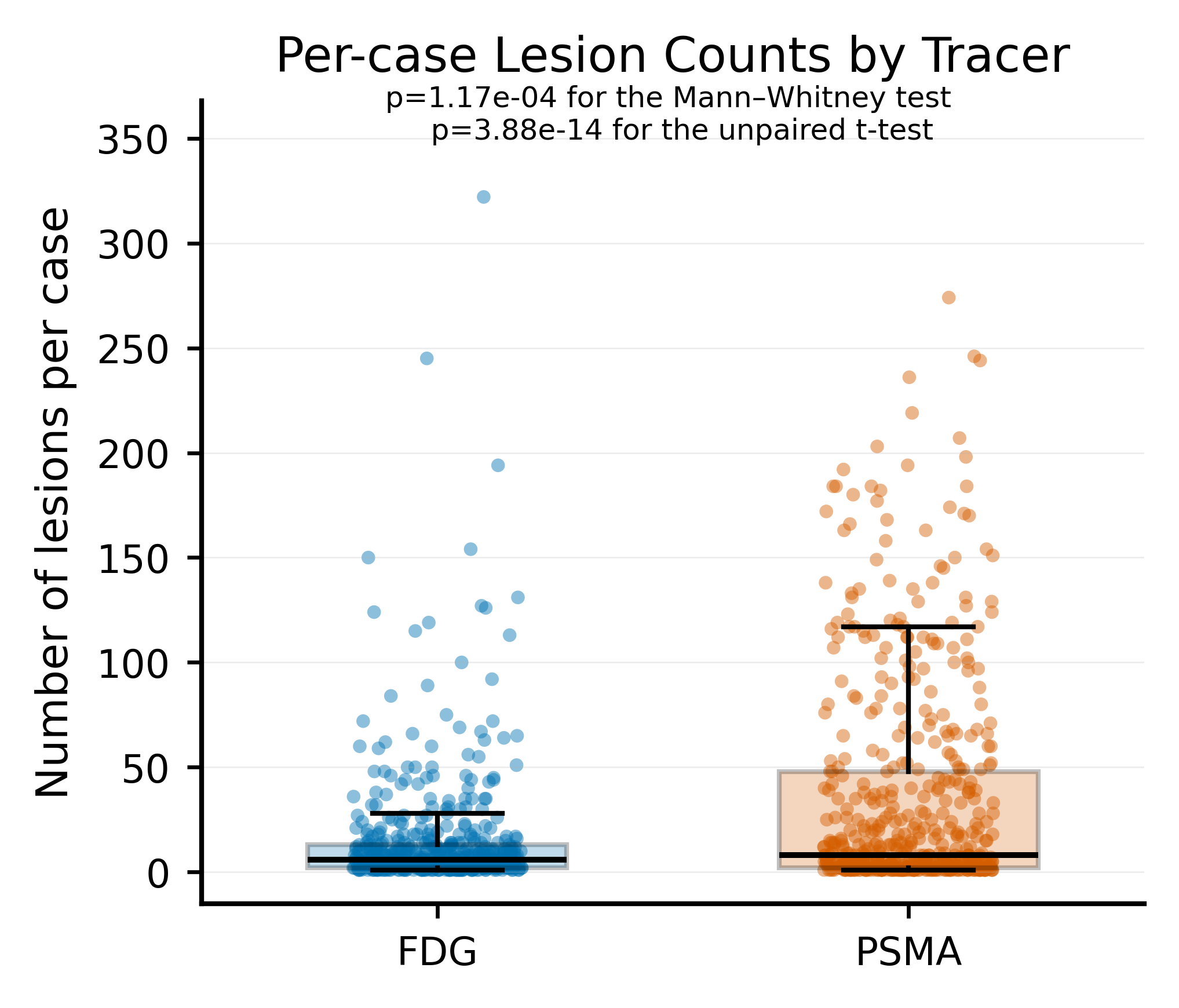}
	\caption{ }
	\label{figs:les_count}
	\end{subfigure}
	\begin{subfigure}[b]{0.45\textwidth}
		\centering
\includegraphics[width=\textwidth]{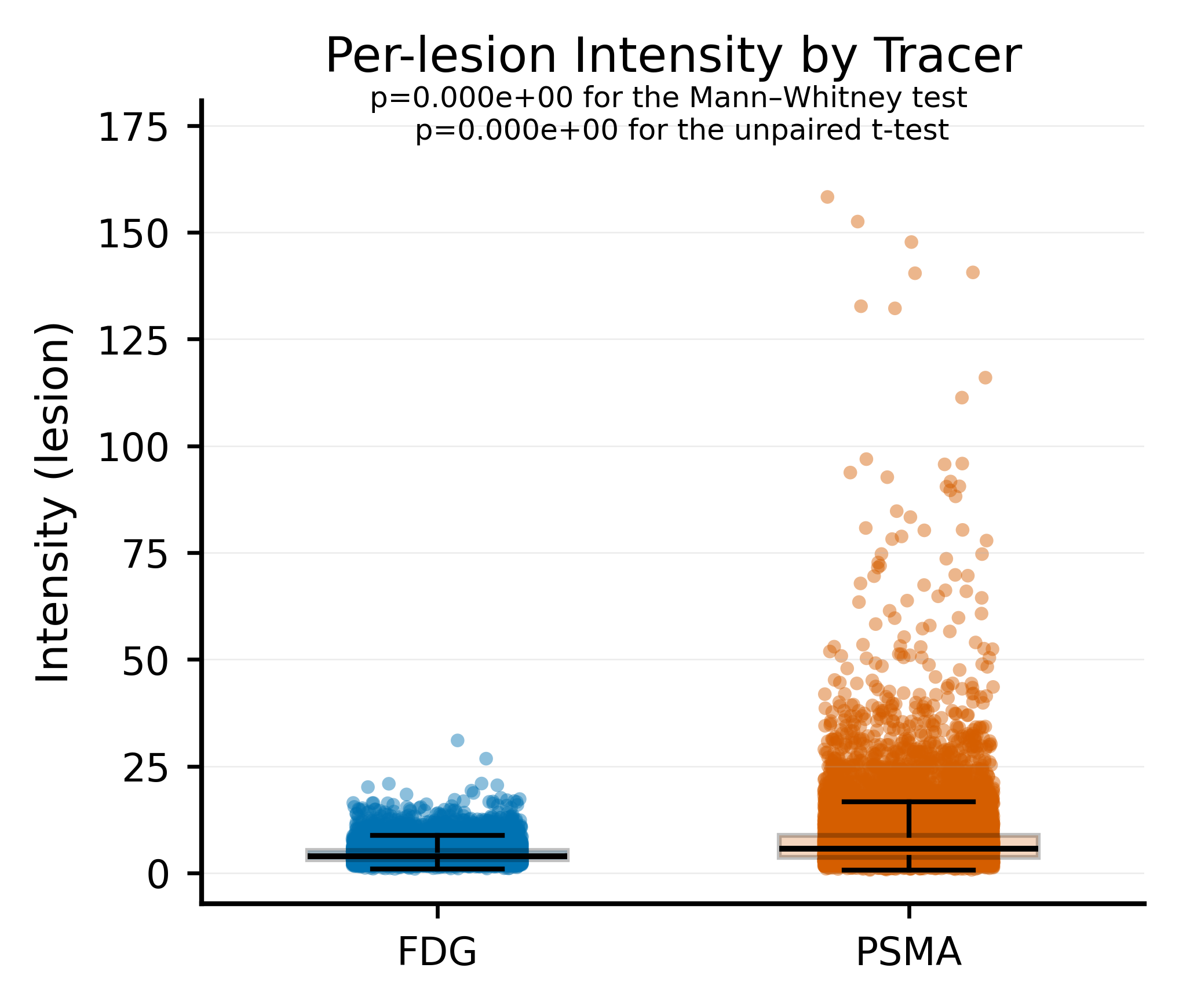}
	\caption{ }
    \label{figs:suvmean}
	\end{subfigure}
    \caption[example]{Tracer-specific variations in the AutoPET-III dataset, where, compared to FDG samples, PSMA samples demonstrate a higher mean and larger variance with statisitcal significant difference ($p \ll 0.05$) in both (a) per-case lesion counts  and (b) per-lesion intensities.}
    \label{figs:stat_tracer}
\end{figure}

\item \textbf{Lack of generalizabillity assessment on unseen datasets for whole-body lesion segmentation:} 
 
Deep learning frameworks have demonstrated strong performance on homogeneous data distributions with carefully designed models, but they often struggle in heterogeneous settings. Such heterogeneity can arise from variations in intensity, resolution, noise characteristics, artifact patterns, anatomical structures, label distributions, and acquisition protocols \cite{yoon2024domain}. As a result, domain generalization has been extensively studied in medical imaging for various organs, including cardiac \cite{chen2022maxstyle}, prostate \cite{liu2020shape}, breast \cite{xu2022improved}, and liver \cite{wen2024denoising}. Similar to these applications, whole-body lesion segmentation models must generalize across diverse domains, including different institutions, tracers, scanners, and disease types, since the limited availability of domain-specific whole-body datasets makes training separate models for each domain impractical. However, to the best of our knowledge, prior work has not systematically evaluated generalization in the context of whole-body lesion segmentation.  
 
\end{enumerate}

In this work, we frame four Research Questions (RQs) encapsulating these shortcomings and address them through Bayesian Deep Learning (BDL) and Uncertainty Quantification (UQ), uncertainty–misclassification correlations and uncertainty-aware modelling. Using the state-of-the-art nnU-Net model \cite{isensee2021nnu} initialized with pretrained weights by Rokuss \textit{et al.} \cite{rokuss2024fdg} as our baseline for whole-body PET/CT lesion segmentation, our key research questions and findings are: 

\begin{itemize}
    \item \textbf{\textit{RQ1: How sensitive are nnUNet-based deterministic whole-body lesion segmentation models to initialization and data batching, and can Bayesian Deep Learning help generate predictions robust to training stochasticity?}} We assess the effect of parameter initialization and mini-batching on lesion segmentation performance and find wide variability in nnU-Net performance even with small changes in parameter initialization and mini-batching. We draw on the extensive literature in Bayesian Deep Learning (BDL) \cite{gal2016dropout, abdar2021review} and adopt the deep ensemble approach \cite{lakshminarayanan2017simple} to develop a Bayesian whole-body segmentation model, in which multiple nnU-Net models with identical architectures and optimization objectives are trained using different parameter initializations and data batch configurations. Instead of relying on a single deterministic version of the model that yields voxel-wise prediction probabilities, we compute the expected values from multiple training iterations of the same model to obtain more reliable, repeatable, and stable predictions.

    In addition to achieving reliable and stable predictions, we use our Bayesian ensembling approach to quantify the total predictive uncertainty, and decompose it into aleatoric (data) and epistemic (model) uncertainty. Aleatoric uncertainty represents inherent data noise that cannot be reduced without acquiring additional or higher-quality information, whereas epistemic uncertainty captures the model’s gaps in knowledge and can be mitigated through further training or data diversification \cite{der2009aleatory}. During inference, our model  generates voxel-wise prediction probability as well as uncertainty maps that capture complementary information about the model's predictions as well as the model's uncertainty in these predictions. %We provide qualitative visualizations of such uncertainty maps in Figures \ref{figs:qual_uncer_fdg} and \ref{figs:qual_uncer_psma}. 

\item \textbf{\textit{RQ2: Is there any relationship between PET/CT lesion prediction uncertainty and prediction misclassification?}} Epistemic uncertainty, as a quantitative indicator of a model’s knowledge gap, is theoretically expected to correlate with prediction misclassification in deep learning systems \cite{lakshminarayanan2017simple}. In this work, we examine whether epistemic uncertainty in whole-body lesion segmentation can reliably identify voxel-wise misclassifications between lesion and non-lesion regions, thereby serving as a proxy for prediction reliability. Using validation samples from the AutoPET-III dataset, we evaluate how effectively uncertainty distinguishes misclassifications such as FPs and FNs. We find that FP voxels show consistently higher epistemic uncertainties compared to FN voxels where epistemic uncertainties are more diverse. This suggests that epistemic uncertainty from the baseline model can be used to distinguish True Positive (TP) from FP voxels more reliably over True Negative (TN) from FN  voxels (Figure \ref{figs:unc_AUROC_AUPRC_nnunet}).

\item \textbf{\textit{RQ3: How can we optimally leverage epistemic uncertainty to improve model performance and prediction reliability?}} Would a model's performance improve if we provide it with its own knowledge gaps (captured by epistemic uncertainty), much like improving human performance by providing them feedback? A previous study by Hartmann \textit{et al.} \cite{hartmann2021bayesian} employed uncertainty maps as additional inputs and achieved improved segmentation performance in Synthetic Aperture Radar (SAR) images to segment glacier regions from water or broken ice regions. Motivated by this, we augment the input channels with epistemic uncertainty maps (that capture model's knowledge gaps) derived from the Bayesian whole-body segmentation model and train an Uncertainty-Augmented model. When compared on the validation set, the two-channel (PET and CT) Bayesian base model exhibits a precision-oriented behavior with lower False Positive Volume (FPVol), whereas the uncertainty-augmented three-channel (PET, CT, uncertainty maps) model favors recall with reduced FNVol. Since the Autopet-III dataset includes multi-tracer cases with wide variability in lesion type, number, location, and size, we train a case-adaptive router that builds on this structured precision–recall trade-off between the Bayesian baseline and uncertainty-augmented models. This case-adaptive router is a shallow convolutional neural network model that takes the PET image of a case as input and predicts whether the case would benefit more from a precision-oriented or recall-oriented model. This early case-adaptive routing strategy leads to improved overall Dice performance on the multi-tracer, pan-cancer cases, compared to using either model individually, demonstrating that learned model selection can effectively exploit the complementary strengths of the Bayesian baseline and uncertainty augmented models.

\item
\textbf{\textit{RQ4: How well does the uncertainty-aware nnU-Net generalize to unseen datasets?}}
Generalization to unseen datasets and external datasets are critical for any medical image segmentation models. 
In this study, we investigate the first three research questions with the training and validation samples of the publicly available AutoPET-III dataset, which comprises multi-tracer, pan-cancer samples. To evaluate generalization, we test our approach on two unseen datasets: (1) an in-distribution split of AutoPET-III, where $321$ scans are held out as a test set, and (2) an out-of-distribution Deep-PSMA dataset \cite{meakin_2025_17701815} consisting of $100$ patients with metastatic prostate cancer, each with both a PSMA-PET/CT and an FDG-PET/CT scans. The Deep PSMA dataset was designed to segment whole body disease burden on PSMA PET/CT (a mix of $^{68}$Ga-PSMA-617 and $^{18}$F-DCFPyL) and FDG PET/CT images for staging of patient for 177 Lu-PSMA Therapy. Although our training set included PSMA-PET/CT for prostate cancer cases, it did not include any FDG PET/CT for prostate cancer. The FDG-PET/CT images in our training set were for other cancers (lung, melanoma, lymphoma) that show different distribution than prostate cancer metastases. Thus, the Deep PSMA dataset presents a unique unseen dataset to evaluate generalizability of our models. Our results show that uncertainty augmentation improves performance on PSMA cases within the unseen AutoPET-III test set, and enhances both FDG and PSMA performance on the Deep-PSMA dataset. 
\end{itemize}

To the best of our knowledge, this work is the first to (a) study how sensitive nnU-Net model is to training stochasticity in multi-tracer, pan-cancer PET/CT datasets, and how Bayesian ensembling can generate improved predidctions, (b) quantify total uncertainty and decompose it to epistemic and aleatoric compoenents for whole-body multi-tracer, pan-cancer lesion segmentation task, (c) leverage the quantified epistemic uncertainty to reduce the models' knowledge gaps, improving performance across different radiotracers in PET/CT imaging, and (d) test the generalizability of uncertainty-augmented models in unseen datasets.  
We are also the first to evaluate how well epistemic uncertainty captures model misclassifications and investigate whether any systematic patterns emerge across two types of misclassifications (false positives and false negatives). 

%We also show that analysing uncertainty–misclassification correlations provides insight into when uncertainty information can be used to reliably distinguish between true and false predictions. %how and when uncertainty augmentation will benefit the final segmentation output.  
%Our contributions in this paper are summarized below.
%\begin{itemize} \itemsep -5pt
 %   \item We propose a novel, multi-tracer, pan-cancer, whole-body lesion segmentation model that: (a) uses Bayesian ensembling of different deterministic nn-UNet models to create a Bayesian nn-UNet model that is robust to stochasticity in training stemming from random initialization and data-batching; (b) quantifies epistemic and aleatoric prediction uncertainties for multi-tracer, pan-cancer whole-body lesion segmentation task; (c) integrates an uncertainty-augmented module to close model's knowledge gaps, improving performance; and (d) integrates a case-adaptive routing module to select the best model for different tracer and cancer type for optimum performance across different cases. 

  %  \item We evaluate how well epistemic uncertainty captures model misclassifications and we investigate whether any systematic patterns emerge across two types of misclassifications.

 %   \item We evaluate the performance of our proposed framework on held-out in-distribution AutoPET-III test samples, as well as on the out-of-distribution Deep PSMA dataset. 
%\end{itemize}

\section{Materials and methods}

\subsection{Problem formulation for whole-body lesion segmentation} \label{secs:Bayes_prob}
Our goal is to automatically identify and segment lesions in a whole-body 3D PET/CT scan. Therefore, we aim to develop a 3D binary segmentation model that examines every voxel in the 3D whole-body scan and classifies it as either lesion or non-lesion. The input space is defined as $\mathcal{X} \subset \Re^{H \times W \times D \times C_{in}}$ where $(H,W,D)$ denote the spatial dimensions (height, width, and depth), and $C_{in}$ corresponds to the number of imaging channels. The output space is a binary segmentation mask $\mathcal{Y} \subset \{0,1\}^{H \times W \times D \times 1 }$ where each voxel $v=\{i,j,k\}$ is labeled as $y_{v}=1$ for being a foreground (lesion), and $y_{v} =0$ othwise (background). Given a training dataset $\mathcal{D}^{train} = \{(X^i \in \mathcal{X},Y^i \in \mathcal{Y})\}_{i=1}^N$, we aim to learn a segmentation model $\theta$ that maps a 4D data $X^i$ to a 3D segmentation mask, as follows 
\begin{equation}\label{eqn:theta}
    \theta: \Re^{H \times W \times D \times C_{in}} \rightarrow \{0,1\}^{H \times W \times D \times 1}
\end{equation}

\subsection{Dataset and pre-processing}\label{sec:dataset}

\textbf{Dataset:} We use two different publicly available datasets, the Autopet-III \cite{Ingrisch2024_autoPETIII} and the Deep-PSMA \cite{meakin_2025_17701815} for this study (Table \ref{tab:data_summary}). We use the AutoPET-III dataset for training, validation, and independent testing, whereas we use the Deep-PSMA dataset exclusively as an out-of-distribution test set to assess the generalizability of our trained models.

\begin{table}[htbp]
\small
\centering
\caption{Dataset summary of AutoPET-III and Deep PSMA with decomposed statistics for FDG and PSMA. M=Male, F=Female, NS=Not Specified, M=Melanoma, Lu=Lung Cancer, Ly=Lymphoma, Neg=Negative, Pr=Prostate, $dz$ = distance, $\mu_{PET} = $ Mean PET Intensity }
\label{tab:data_summary}
\begin{tabular}{@{}lcccc}
\hline
\multirow[]{2}{*}{Properties} & \multicolumn{2}{c}{AutoPET-III} & \multicolumn{2}{c}{Deep-PSMA} \\
\cmidrule(r){2-3} \cmidrule(r){4-5}
& FDG & PSMA & FDG & PSMA\\ 
\hline
$\#$ Patients  & 1014 & 597 & 100 & 100  \\
Sex (M/F/NS) & 570 / 440/ 4 & 597/ 0 & 100/ 0 & 100/0 \\
Age (years) & 59.4 ± 16.0 (11--95)  & 71.43 ± 8.18 (48--92) & N/A & N/A \\
Data  & PET-CT & PET-CT & PET-CT & PET-CT  \\ 
\hline
Disease dist. & {\begin{tabular}{c}  M/Lu/Ly/Neg  \\ 188/168/145/513 \end{tabular}}  & {\begin{tabular}{c} Pr/Neg\\ 537/60\end{tabular}} & {\begin{tabular}{c}Pr/Neg\\100/0\end{tabular}} &{\begin{tabular}{c} Pr/Neg\\100/0 \end{tabular}}\\
\hline
$\#$lesions (per case) & 15.15 $\pm$ 29.17 & 35.04 $\pm$ 51.70 & 38.38 $\pm$ 39.66 & 68.41 $\pm$ 58.35 \\

$\mu_{PET}$ (per lesion) & 4.56 $\pm$ 2.32 & 7.48 $\pm$ 7.12 & 3.73 $\pm$ 1.19 & 5.64 ± 3.33 \\

$\#$ slices (per vol) &  200--661 & 135--963 & 199--1206 & 195--1261   \\

In-plane res. (mm) & 2.04--2.04 & 2.73--4.07 & 2.73--5.47 & 1.59--5.47  \\

$dz$ bet. slices (mm) & 3.00--3.00 & 2.00--5.00 & 1.5--5.00 &  1.5--5.00 \\
train/val/test & 664/151/199 & 379/96/122 &0/0/100 & 0/0/100 \\
\hline
\end{tabular}%
\end{table}

The AutoPET-III dataset \cite{Ingrisch2024_autoPETIII} contains a total of $1,611$ PET/CT studies: $1,014$ acquired with FDG tracers and $597$ with PSMA tracers. Among the FDG cohort, $513$ cases are lesion-free, while the remaining $501$ include $188$ malignant melanoma cases, $168$ lung cancer cases, and $145$ lymphoma cases. Within the PSMA cohort, $60$ cases are lesion-free and the remaining $537$ represent prostate cancer patients with metastatic disease. For cancerous cases, the number and location of the lesions vary significantly, ranging from 1 to $294$ for PSMA cases (prostate cancer metastasis), and to $1031$ for FDG cases. The PET intensity values also showed significant variability, with average intensity values within lesions ranging from $0.99$ to $31.06$ for FDG cases, and $0.74$ to $158.29$ for PSMA cases.  

In AutoPET-III, FDG cases were reviewed by a radiologist and nuclear medicine physician to identify primary tumors and metastases. Malignant FDG-avid lesions were manually segmented slice by slice on PET images \cite{gatidis2022whole}. For PSMA cases, PSMA-avid primary and metastatic lesions were delineated by a reader with three years of hybrid imaging experience. Candidate lesions were pre-segmented using user-defined uptake thresholds and manually refined slice by slice to generate 3D binary masks \cite{jeblick2024psmapetctlesions}.

The Deep-PSMA dataset \cite{meakin_2025_17701815} consists of $100$ PET/CT patients, each with an FDG and PSMA PET/CT scan acquired for staging prior to LuPSMA therapy. All patients had metastatic prostate cancer. Although each patient had both FDG and PSMA scan, there was variability in the visibility of lesions on the two tracers. The number of lesions varies considerably across patients, ranging from $1$ to $149$ in FDG cases and from $2$ to $235$ in PSMA cases. The average PET intensity values within lesions range from $1.97$ to $19.38$ for FDG cases and from $2.99$ to $50.77$ for PSMA cases. It may be noted that the AutoPET-III dataset does not contain any FDG prostate cancer metastasis cases, which makes this dataset a unique out of distribution test set to evaluate.

In the Deep-PSMA dataset, candidate regions were identified using PET SUV thresholds: SUV $>3$ for PSMA scans and liver mean plus two standard deviations for FDG scans. Expert nuclear medicine physicians with at least five years of specialization then reviewed these regions and labeled tracer uptake as lesion or physiological/non-lesion.

\textbf{Data Preprocessing:} We follow the methodology of Rokus \textit{et al.} \cite{rokuss2024fdg}. Specifically, we remove intensity values below the 5th and above the 95th percentile for both PET and CT scans, and apply Z-score normalization to the remaining voxel intensities. The mean voxel spacing is computed as $(x = 2.04, y = 2.04, z = 3.0)$, and all PET and CT volumes are resampled to this uniform resolution. During training, random cropping is performed with a patch size of $[192, 192, 192]$, while inference is conducted using a sliding-window approach consistent with the standard nnU-Net protocol \cite{isensee2021nnu}. We construct a held-out test set of $321$ AutoPET-III images ($199$ FDG and $122$ PSMA), using the remaining $1{,}043$ images for training and $247$ for validation. To prevent data leakage, splits are defined so that longitudinal scans from the same patient appear in only one of the train, validation, or test sets. All samples from the Deep-PSMA dataset are used exclusively as a held-out test set and are preprocessed using the same pipeline as AutoPET-III.

% \textcolor{red}{maybe add the train/val/test split here? you can add a line to the dataset table too}

\subsection{Overview of our proposed method}

\begin{figure} [ht]
   \begin{center}
   \includegraphics[width=1\textwidth]{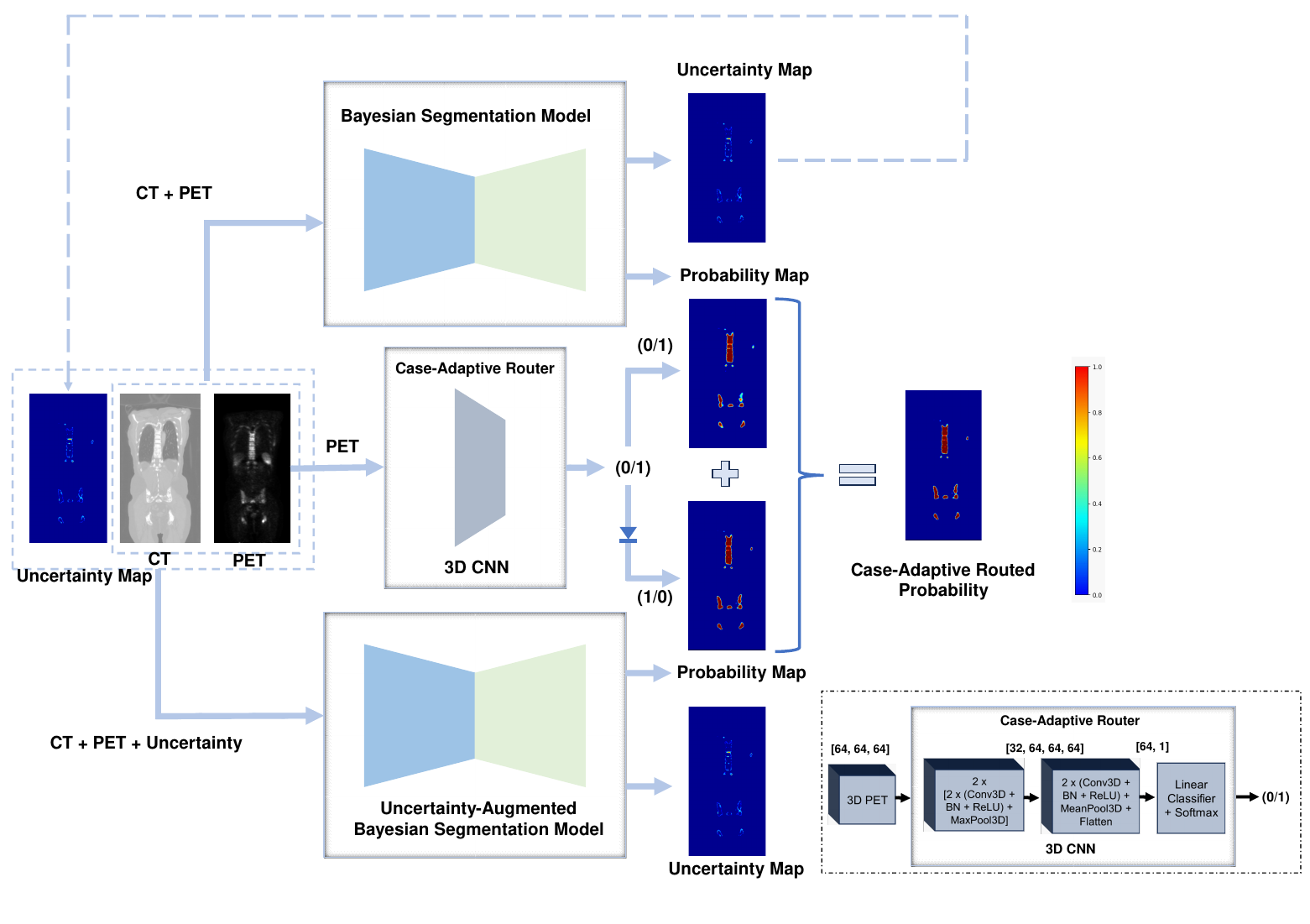}
   \caption[example]{Overview of our proposed approach.}
   \label{figs:overview}
	\end{center}
\end{figure}

An overview of the proposed framework is presented in Figure \ref{figs:overview}, which consists of four key modules: (i) the Bayesian whole-body segmentation model, (ii) uncertainty quantification, (iii) the uncertainty-augmented segmentation model, and (iv) the case-adaptive Routing block. Section \ref{sec:bayes_seg} describes the Bayesian modelling for whole-body lesion segmentation. Section \ref{sec:uq} introduces the fundamentals of uncertainty quantification and details how uncertainty maps are obtained from the Bayesian model. Section \ref{sec:uab} then explains how these uncertainty maps are incorporated as an additional input channel, expanding the model from two input modalities (CT and PET) to three (CT, PET, and uncertainty). Finally, Section \ref{sec:tracer_aware} outlines the case-adaptive routing network, which is trained solely on PET images and dice difference of base and augmented models to predict the best model for each case.  Each of these four components is described in detail in the following subsections. 

\subsection{Bayesian whole-body lesion segmentation} 
\label{sec:bayes_seg}

State-of-the-art whole-body lesion segmentation models are typically deterministic, where the network parameters $\theta$ are optimized to a single point estimate $\theta^*$ using maximum likelihood estimation on the training dataset $\mathcal{D}^{train}$. In the following sections, we first describe our deterministic nnU-Net baseline and then extend it to Bayesian prediction using deep ensembling.

\subsubsection{Deterministic whole-body lesion segmentation models} 
\label{secs:det_base}

We use nnU-Net \cite{isensee2021nnu} as the deterministic baseline for whole-body lesion segmentation. Following the AutoPET-III training strategy of Rokuss \textit{et al.} \cite{rokuss2024fdg}, one of the top-performing challenge methods, the input consists of CT and PET channels, i.e., $C_{in}=2$. The model parameters are optimized using an equally weighted combination of Dice and cross-entropy losses:
\begin{equation} \label{eqn:det_loss}
    \theta^* = \arg \min_{\theta} \sum_{i=1}^{N} 
    \left( w_{DICE}\mathcal{L}_{DICE} + w_{CE}\mathcal{L}_{CE} \right),
\end{equation}
where $w_{DICE}=w_{CE}=0.5$, and $N$ denotes the number of training samples. We train each model for up to 120 epochs with early stopping if validation performance does not improve for 15 epochs. This shorter schedule, compared with the 1500 epochs used in \cite{rokuss2024fdg}, enables computationally feasible analysis of model sensitivity and deep ensembling. At inference, the trained model predicts the voxel-wise lesion label for a test image $X^{test}$ as
\begin{equation}\label{eqs:det}
    \bar{y}_v = \arg \max_{\{0,1\}} p(y_v|X^{test},\theta^*), 
    \quad v=(i,j,k).
\end{equation}

\subsubsection{Posterior probability through Bayesian modelling} 
\label{sec:prob_base}

The probability in Eqn. (\ref{eqs:det}) is obtained from the softmax output of a single deterministic model. However, such point-estimate predictions are limited for uncertainty estimation \cite{lakshminarayanan2017simple}. We therefore adopt a Bayesian formulation in which the model parameters $\theta$ are treated as random variables. The posterior predictive probability for each voxel is given by
\begin{equation}\label{eqs:post}
    p(y_v | X^{test}, \mathcal{D}^{train}) =
    \int_{\theta} P(\theta |\mathcal{D}^{train}) 
    P(y_v|X^{test},\theta) d\theta
    =
    E_{\theta \sim P(\theta |\mathcal{D}^{train})}
    \left[P(y_v|X^{test},\theta)\right]
\end{equation}

Since exact posterior inference is intractable for deep networks, we approximate this expectation using deep ensemble averaging \cite{lakshminarayanan2017simple}:
\begin{equation}\label{eqs:post_approx}
    p(y_v | X^{test}, \mathcal{D}^{train}) 
    \approx 
    \frac{1}{K} \sum_{k=1}^{K} 
    P(y_v|X^{test},\theta^*_k),
\end{equation}
where $\theta^*_k$ denotes the parameters of the $k^{th}$ independently trained ensemble member.

\subsection{Uncertainty quantification}
\label{sec:uq}

The posterior predictive probability in Eqn. (\ref{eqs:post}) enables voxel-wise uncertainty estimation. We define total uncertainty $\mathcal{U}_{tot}$ using predictive entropy, which can be decomposed into aleatoric and epistemic components \cite{kendall2017uncertainties,depeweg2018decomposition}:
\begin{equation}\label{eqs:epi_image}
    \underbrace{\mathcal{H}\left(P(y_v | X^{test}, \mathcal{D}^{train})\right)}
    _{\text{Total Unc. } \mathcal{U}_{tot}}
    =
    \underbrace{
    E_{\theta \sim P(\theta |\mathcal{D}^{train})}
    \left[
    \mathcal{H}\left(P(y_v | X^{test}, \theta)\right)
    \right]}
    _{\text{Aleatoric Unc. } \mathcal{U}_{ale}}
    +
    \underbrace{
    \mathcal{I}\left[y_v,\theta | X^{test}, \mathcal{D}^{train}\right]}
    _{\text{Epistemic Unc. } \mathcal{U}_{epi}}.
\end{equation}

The aleatoric term captures expected prediction uncertainty due to inherent data ambiguity, while the epistemic term measures model disagreement induced by uncertainty over $\theta$. Thus, epistemic uncertainty increases when test samples are poorly represented by the training distribution \cite{gal2016uncertainty}. Using the deep ensemble approximation, we estimate these quantities as
\begin{equation}\label{eqs:uncer}
    \begin{split}
        \mathcal{U}_{tot} 
        &\approx 
        \mathcal{H}\left(
        \frac{1}{K}\sum_{k=1}^{K} 
        P(y_v|X^{test},\theta^*_k)
        \right), \\
        \mathcal{U}_{ale} 
        &\approx 
        \frac{1}{K}\sum_{k=1}^{K}
        \mathcal{H}\left(
        P(y_v|X^{test},\theta^*_k)
        \right), \\
        \mathcal{U}_{epi} 
        &= 
        \mathcal{U}_{tot} - \mathcal{U}_{ale}.
    \end{split}
\end{equation}

The resulting $\mathcal{U}_{tot}$, $\mathcal{U}_{ale}$, and $\mathcal{U}_{epi}$ have the same spatial dimensions as the predicted segmentation mask, $\{H \times W \times D \times 1\}$, and are therefore represented as voxel-wise uncertainty maps.

\subsubsection{Relationship between epistemic uncertainty and misclassification}\label{sec:cor_epi}

As discussed in Section \ref{sec:uq}, epistemic uncertainty reflects the model’s knowledge gap; consequently, voxels with high epistemic uncertainty are expected to correspond to misclassified predictions \cite{lakshminarayanan2017simple}. Building on this idea, Malinin \textit{et al.} \cite{malinin2018predictive} proposed using epistemic uncertainty to identify misclassified samples. Motivated by this, we conduct a study with the validation set samples (Section \ref{sec:anal_qual}) to investigate the relationship between epistemic uncertainty and misclassification.  %Our analysis shows that epistemic uncertainty maps from the Bayesian base model while correctly identifying false positives from true positives, fail to distinguish false negatives from true negatives consistently. %Our analysis shows that epistemic uncertainty maps provide informative signals for identifying missed voxels (false negatives) and may capture complementary information that is not fully represented in the input PET/CT images. To further assess this complementary information, we introduce an uncertainty-augmented model, detailed in the following section.  

\subsection{Uncertainty-augmented segmentation model}\label{sec:uab}

We design an uncertainty-augmented model that leverages the epistemic uncertainties estimated by the Bayesian model as a means to incorporate model's knowledge gaps. Specifically, we extract epistemic uncertainty maps for all samples in the AutoPET-III dataset and incorporate them as an additional input channel to a deterministic segmentation framework. The model is then trained using the same loss function as in Eqn. \ref{eqn:det_loss}, resulting in a three-channel deterministic segmentation model. To clearly delineate the distinction between the models, we summarise their parameter configurations as follows 

\begin{equation}
    \begin{split}
    &\theta: \Re^{H \times W \times D \times C_{in} = 2} \rightarrow \{0,1\}^{H \times W \times D \times 1} \ \ \ \text {$C_{in} = \|\{CT, PET\}\|$} \\
      &\theta_{UA}: \Re^{H \times W \times D \times C_{in} = 3} \rightarrow \{0,1\}^{H \times W \times D \times 1} \ \ \ \text {$C_{in} = \|\{CT, PET,\mathcal{U}_{epi}\}\|$}   \\
       % &p_{UAB}(y_v | X^{test}, \mathcal{D}^{train}, \mathcal{U}^{train}_{epi,BB}) = \frac{1}{K} \sum_{k}^K  P(y_v|  X^{test}, \mathcal{U}^{test}_{epi,BB},\theta_{UAB,k}^*) \\
       % &\mathcal{U}_{epi,UAB} = \mathcal{I}[P(y_v, \theta_{UAB} | X^{test}, \mathcal{U}^{test}_{epi,BB},\mathcal{D}^{train}, \mathcal{U}^{train}_{epi})] =  \mathcal{U}_{tot,UAB} - \mathcal{U}_{ale,UAB}
    \end{split}
\end{equation}

Analogous to the two-channel Bayesian model derived from the deterministic parameters $\theta_k$ in Eqn. \ref{eqs:post_approx}, we construct a Bayesian version of the uncertainty-augmented model. The deterministic members of this model are trained for 90 epochs, compared to 120 epochs used for the base model. Mathematically, the posterior probability of a voxel being a lesion or not from the Bayesian uncertainty-augmented model can be expressed as follows 

\begin{equation}\label{eqs:post_approx_aug}
    p_{UA}(y_v | X^{test}, \mathcal{D}^{train}) \approx \frac{1}{K} \sum_{k}^K  P_{UA}(y_v|  X^{test}, \theta^*_{aug,k})
\end{equation}

\subsection{Case adaptive routing}\label{sec:tracer_aware}

On examining the performance of the Bayesian baseline and the Bayesian uncertainty-augmented models on the validation set samples (Sect. \ref{sec:abla_carbu}, Figure \ref{figs:comple_fpvol_fnvol}), we find a clear and statistically significant disparity in misclassification characteristics: the uncertainty-augmented model achieves significantly lower FNVol, whereas the base model yields significantly FPVol. This indicates a structured precision–recall trade-off between the two models that is not captured by Dice alone. Since the AutoPET-III dataset includes a wide variability in lesion and PET intensity distribution, we hypothesize that an automated early routing module to adaptively select the optimal model for each PET-CT case would enable a precision-recall balance. Specifically, we train a case-adaptive router using PET as the input signal to predict which model is expected to perform better. We explore two training objectives for the router: Mean Squared Error (MSE) and Cross Entropy (CE) loss. Under the MSE formulation, the router is trained to regress the Dice difference between the base and augmented models. During inference, a positive predicted difference routes the case to the base model, while a negative prediction selects the uncertainty-augmented model. Under the CE formulation, we convert the Dice difference into binary labels, assigning label ‘0’ when the base model outperforms the augmented model and ‘1’ otherwise. For each test sample, the trained router predicts the preferred model, and the sample is subsequently processed by either the base or the uncertainty-augmented model. Mathematically, given a test sample $X^{\text{test}}$, the MSE-based router predicts a continuous performance-difference score
\begin{equation}
\Delta_\theta\!\left(X_{\text{PET}}^{\text{test}}\right) \in \mathbb{R},
\end{equation}
where the router is trained to regress the Dice difference between the base and uncertainty-augmented models:
\begin{equation}
\Delta^{\text{test}} =
\mathrm{Dice}\!\left(f_b(X^{\text{test}}), Y^{\text{test}}\right)
-
\mathrm{Dice}\!\left(f_a(X^{\text{test}}), Y^{\text{test}}\right).
\end{equation}
Thus, a positive predicted difference indicates that the base model is expected to outperform the uncertainty-augmented model, whereas a negative predicted difference indicates that the uncertainty-augmented model is expected to perform better. The routing decision is given by
\begin{equation}
\hat{z}^{\text{test}} =
\begin{cases}
0, & \text{if } \Delta_\theta\!\left(X_{\text{PET}}^{\text{test}}\right) > 0, \\
1, & \text{if } \Delta_\theta\!\left(X_{\text{PET}}^{\text{test}}\right) \leq 0,
\end{cases}
\end{equation}
where $\hat{z}^{\text{test}}=0$ and $\hat{z}^{\text{test}}=1$ correspond to selecting the base and uncertainty-augmented models, respectively. The final segmentation is then obtained as
\begin{equation}
\hat{Y}^{\text{test}} =
\begin{cases}
f_b\!\left(X^{\text{test}}\right), & \text{if } \hat{z}^{\text{test}} = 0, \\
f_a\!\left(X^{\text{test}}\right), & \text{if } \hat{z}^{\text{test}} = 1.
\end{cases}
\end{equation}

% Mathematically, given a test sample $X^{\text{test}}$, the CE-based router predicts class probabilities
% \begin{equation}
% p_\theta(c \mid X_{\text{PET}}^{\text{test}}), \quad c \in \{0,1\},
% \end{equation}
% where $c=0$ and $c=1$ correspond to selecting the base and augmented models, respectively.
% The routing decision is given by
% \begin{equation}
% \hat{z}^{\text{test}} =
% \arg\max_{c \in \{0,1\}} p_\theta(c \mid X_{\text{PET}}^{\text{test}}).
% \end{equation}
% The final segmentation is then obtained as
% \begin{equation}
% \hat{Y}^{\text{test}} =
% \begin{cases}
% f_b\!\left(X^{\text{test}}\right), & \text{if } \hat{z}^{\text{test}} = 0, \\
% f_a\!\left(X^{\text{test}}\right), & \text{if } \hat{z}^{\text{test}} = 1.
% \end{cases}
% \end{equation} 

% We also observe in Figure \ref{figs:comple_fdg_psma} that there exists a weak, tracer-dependent trend in Dice performance between FDG and PSMA across the base model and the uncertainty-augmented model. However, these differences are not statistically significant, indicating limited complementarity at the input level. 

\section{Ablation experiments}\label{sec:exper}

Our model design choices were based on ablation experiments to assess:

\begin{enumerate}
\item how sensitive nnUNet-based deterministic whole-body cancer segmentation models are to intialization and data-batching, and if Bayesian ensembling (Section \ref{sec:bayes_seg}) provides a more consistent whole-body lesion segmentation performance;

\item how to quantify and decompose prediction uncertainty (Section \ref{sec:uq}), and how to understand the relationship between epistemic uncertainty and model errors;

\item how to levearge epistemic uncertainty to improve model performance in pan-cancer multi-tracer whole body PET/CT datasets
\end{enumerate}
\paragraph{Summary of ablation experiments:} We utilise 1043 training and 247 validation samples from AutoPET-III for all of our ablation experiments. We consider a baseline nnU-Net with PET–CT inputs under both deterministic and Bayesian (Base) settings, and an uncertainty-augmented variant with PET–CT–uncertainty inputs under deterministic and Bayesian (Aug) settings. To exploit complementary strengths of the base and uncertainty-augmented models, we investigate combination strategies through late fusion (intersection, average probability, maximum probability) and early routing approaches. For early routing, we evaluate both a tracer-aware routing and a learned router approach. In learning the router, we investigate multiple design choices, including loss functions (vanilla cross-entropy, vanilla MSE, and signaware MSE), input modalities (PET, CT, PET-uncertainty, and PET–CT), and different input reshape sizes (64×64×64, 80×80×80, and 120×120×120). We provide an overview of all experimental configurations evaluated in this study in Table \ref{tab:ablation_experiments}. 

\begin{table}[htbp]
\caption{Overview of all experimental configurations evaluated in this study. We utilize both deterministic and Bayesian ensembling of baseline and uncertainty augmented models. Afterwards, we investigate complementary strengths of base and augmented model through late fusion and early routing strategies.}
\label{tab:ablation_experiments}
\begin{center}
\resizebox{\textwidth}{!}{
     \begin{tabular}{|c|c|c|c|c|c|}
     \hline
   \textbf{Model} & \textbf{Inputs/Modes} & \multicolumn{4}{|c|}{\textbf{Method}}  \\
      \hline
      \multirow[]{2}{*}{Baseline nnUNet \cite{rokuss2024fdg} } & \multirow[]{2}{*}{PET-CT} & \multicolumn{4}{|c|}{Deterministic} \\
      & & \multicolumn{4}{|c|}{Bayesian (Base)}\\
      \hline
      \multirow[]{2}{*}{Uncertainty Augmented nnUNet } & \multirow[]{2}{*}{PET-CT-Uncertainty} & \multicolumn{4}{|c|}{Deterministic}\\
      & & \multicolumn{4}{|c|}{Bayesian (Aug)}\\
      \hline
      \multirow[]{13}{*}{Combination of Base and Aug} & \multirow[]{3}{*} {Late Fusion} & \multicolumn{4}{|c|}{Intersection} \\ 
      &  & \multicolumn{4}{|c|}{Average Probabiliy} \\
      &  & \multicolumn{4}{|c|}{Maximum Probability} \\
      \cline{2-6}
      &  \multirow[]{10}{*}{Early routing} & \multicolumn{4}{|c|}{Tracer-aware rule} \\
      \cline{3-6}
      & & \multicolumn{4}{|c|}{Learnt Router}\\
      \cline{3-6}
      & &  Loss & Variant & Input & Reshape Size \\
      \cline{3-6}
      & & \multirow[]{3}{*}{CE} & \multirow[]{3}{*}{Vanilla} & \multirow[]{3}{*}{PET} & 64x64x64 \\ 
      &  & & & & 80x80x80 \\
      &  & & & & 120x120x120 \\
      \cline{3-6}
      & & \multirow[]{5}{*}{MSE} & Vanilla & PET & 64x64x64 \\
      \cline{4-6}
      &  &  & \multirow[]{4}{*}{Weighted Signaware} & PET & \multirow[]{4}{*}{64x64x64} \\
      &  &  & & CT & 
       \\
       &  &  &  & PET-Uncertainty & \\
       & & &  & PET-CT & \\
     \hline
    \end{tabular}
    }
\end{center}
\end{table}

% \textcolor{red}{Evaluation methods:...add.. please follow the corrfabr paper structure.. break up the ablation experiments as per bullets above to just describe how you do them first. then add another section of experimental results.. within that, a subsection of results from ablation experiments, and then assessing generalizability on unseen data}

\subsection{Sensitivity analysis of deterministic nnU-Net and Bayesian ensembling}\label{sec:bayes_abla_eval}
As described in Sect. \ref{sec:intro}, deep models can converge to different local optima due to initialization and batching variability, and we forumulate \textbf{RQ1} to address such stochasticity. We follow Rokuss \textit{et al.} \cite{rokuss2024fdg} and train $K=5$ nnU-Net models with identical architectures on the same AutoPET-III splits for 120 epochs. We then apply Bayesian ensemble averaging (Eqn. \ref{eqs:post_approx}) and evaluate whether the ensemble improves over the average performance of individual deterministic models.

\subsubsection{Evaluation} We deploy the well-established segmentation evaluation DICE, False Positive Volume (FPVol), and False Negative Volume (FNVol), similar to Rokuss \textit{et al.} \cite{rokuss2024fdg} and present both the aggregate evaluation (ALL) and decomposed evaluation with FDG and PSMA.

\subsection{Uncertainty quantification and its relationship with misclassifications}\label{sec:anal_qual}

In this section, we address \textbf{RQ2} through quantifying uncertainty and investigating its relationship with misclassification. We first obtain $K$ voxel-wise segmentation probability maps from the $K$ deterministic models $\theta_k^*$. Following Eq. \ref{eqs:uncer}, we compute the entropy from the expected probability as the total uncertainty and the expectation of the individual entropies as the aleatoric uncertainty. The epistemic uncertainty is then derived by subtracting the aleatoric component from the total uncertainty.

\subsubsection{Evaluation}

We evaluate how uncertainty is related to the misclassified voxels, \textit{i.e.} FP and FN voxels, inspired by previous studies \cite{jungo2019assessing, wang2019aleatoric,schott2025uncertainty}. To measure such relationships, we resort to two different types of analysis: 
\begin{itemize}
    \item For each case, we compute the median epistemic uncertainty for TN, TP, FN, and FP voxels. We visualize these distributions using violin plots and evaluate whether the differences are statistically significant. Pairwise statistical comparisons were performed using the Wilcoxon signed-rank test between TN and TP, TP and FN, and FN and FP.   
    
    \item To assess whether epistemic uncertainty distinguishes correctly classified voxels from misclassified ones, we compute voxel-wise Area Under Receiver Operativing Curve (AUROC) on AutoPET-III validation samples, treating misclassified voxels, false positives (FPs) or false negatives (FNs), as \textit{True} and correctly classified voxels as \textit{False}. The analysis is restricted to the union of predicted and ground-truth voxels: non-overlapping predictions are FPs, non-overlapping ground-truth voxels are FNs, and overlapping voxels are true positives. By sweeping the epistemic uncertainty threshold, we generate ROC curves for each case and compute AUROC as a measure of uncertainty-based misclassification detection. Higher AUROC indicates stronger separation between uncertain misclassified voxels and confident correct predictions. Finally, we compare AUROC distributions for FP and FN detection across validation samples using unpaired Welch’s t-test and the Mann–Whitney U test.
\end{itemize}

\subsection{Leveraging uncertainty to improve model performance}\label{sec:abla_aug}

In this ablation study, we investigate \textbf{RQ3} to leverage epistemic uncertainty to improve model performance and prediction reliability. 

\subsubsection{Uncertainty-augmented Bayesian nnU-Net}

We incorporate epistemic uncertainty maps into the segmentation model as an additional input modality, alongside PET and CT, as described in Sect. \ref{sec:uab}. By encoding information related to both false-positive and false-negative regions, these maps provide complementary cues that enable the uncertainty-augmented model to improve lesion detection compared to the base model. 

\paragraph{Evaluation:} We follow the same protocol used for the base Bayesian model in Sect. \ref{sec:bayes_abla_eval}.  

\subsubsection{Case-adaptive routing}
To evaluate \textbf{RQ3}, we first compare FPVol and FNVol between the base and uncertainty-augmented models on validation samples and observe the augmented model consistently reduces FNVol, whereas the base model yields lower FPVol, indicating complementary strengths (Figure \ref{figs:comple_fpvol_fnvol}). Motivated by this, we conduct an ablation study combining the two models through late fusion and early routing strategies. Late fusion includes maximum probability, average probability, and intersection-based fusion. For early routing, we first use a tracer-based rule that assigns FDG cases to the base model and PSMA cases to the augmented model. We then evaluate learned routers trained with CE and MSE objectives, varying PET input resolution for CE-based routing and testing standard/sign-aware MSE with PET, CT, PET-uncertainty, and PET-CT inputs. 

\paragraph{Training details for case-adaptive router:} The objective of the routing mechanism is to select the most suitable model for a given input. To this end, we use the Dice difference between the base and augmented models as the supervision signal for training the router. In the CE formulation, the target label is set to 0 if the Dice difference is greater than zero (favoring the base model) and 1 if it is less than zero (favoring the augmented model). In the MSE formulation, the router directly predicts the Dice difference; a positive prediction selects the base model, while a negative prediction selects the augmented model. During training and validation of the case-adaptive router, we exclude cases where the absolute Dice difference is below a predefined margin of 0.01 to avoid ambiguous supervision. 
% During testing on unseen data, if the predicted Dice difference is less than 0.01, we apply the tracer-based rule to decide the best model, i.e. FDG samples are passed to

\paragraph{Evaluation:}

We follow the same protocol as the base Bayesian model in Sect. \ref{sec:bayes_abla_eval}, extending it to evaluate various late-fusion and early-routing strategies, including tracer-aware and learned approaches.

\section{Experimental results}
\textbf{Summary results:} We observe that deterministic models exhibit substantial variability in their predictions as reflected in Dice, FPVol, and FNVol (Figure \ref{figs:qual_bayes_fdg_nnunet}, Table \ref{tab:scores_nnunet_bayes}). In contrast, Bayesian ensembling provides more stable and robust performance across all three metrics compared to the average of the deterministic models (Table \ref{tab:scores_nnunet_bayes}). We observe that per-case median epistemic uncertainty is higher for misclassified voxels than for correctly classified ones (Figure \ref{figs:tn_tp_fn_tp_dist_nnunet}). Furthermore, uncertainty is more effective at identifying FP voxels than FN voxels, as indicated by higher AUROC values for FP detection (Figure \ref{figs:AUROC_FN_FP_nnunet}).  The uncertainty-augmented models achieve higher recall with reduced FNVol, but at the cost of lower precision and increased FPVol, relative to the two-channel base Bayesian model for both validation samples (Figure \ref{figs:comple_fpvol_fnvol}, Table \ref{tab:scores_tambu_nnunet_val}) and held-out testing samples (Table \ref{tab:scores_carbu_nnunet_unseen}) from AutoPET-III. The proposed routing approach helps select the optimal of the base and augmented model in the held-out AutoPET-III test set, achieving the best Dice by balancing precision and recall (Tables \ref{tab:scores_tambu_nnunet_val} and \ref{tab:scores_carbu_nnunet_unseen}). For the unseen Deep-PSMA test set, the uncertainty-augmented model improves performance over baseline for all cases (Table \ref{tab:scores_carbu_nnunet_unseen}).

\begin{table}[htbp]
\caption{Bayesian ensembling consistently outperforms the average performance of deterministic models for both the base (2-input: PET and CT) and uncertainty-augmented (3-input: PET, CT, and epistemic uncertainty) settings with validation samples ($N=247$). It achieves higher Dice scores along with lower FPVol and FNVol across ALL, FDG, and PSMA cases. On average, deterministic models also benefit from uncertainty augmentation. Notably, the base Bayesian model attains the highest Dice for FDG cases, whereas the augmented Bayesian model achieves the best Dice for PSMA cases. Additionally, uncertainty augmentation reduces FNVol at the cost of increased FPVol, reflecting a precision–recall trade-off.   }
\label{tab:scores_nnunet_bayes}
\begin{center}
\resizebox{\textwidth}{!}{
     \begin{tabular}{ccllllllllllll}
     \toprule
    \multirow[]{3}{*}{\textbf{Inputs}} & \multirow[]{3}{*}{\textbf{Method}}  & 
    \multicolumn{9}{c}{\begin{tabular}{c} Cancer (N=161) \\ FDG=77, PSMA=84
    \end{tabular}} & \multicolumn{3}{c}{\begin{tabular}{c}No-disease (N=86) \\ FDG=74, PSMA=12\end{tabular}}\\
    \cmidrule(r){3-11} \cmidrule(r){12-14} 
    & & \multicolumn{3}{c}{DICE $\uparrow$} & \multicolumn{3}{c}{FPvol $\downarrow$ } & \multicolumn{3}{c}{FNvol $\downarrow$ } & \multicolumn{3}{c}{FPVol (No-disease) $\downarrow$ }  \\
     \cmidrule(r){3-5} \cmidrule(r){6-8} \cmidrule(r){9-11} \cmidrule(r){12-14}
      & & All & FDG & PSMA & All & FDG & PSMA & All & FDG & PSMA & All & FDG & PSMA    \\
      \midrule
       \multirow[]{7}{*}{\begin{tabular}{c}  PET \\$+$ \\ CT \end{tabular}}& $Det^1$ & 61.1 & 68.5 & 54.4 & 9.9 & 4.8 & 14.5 & 14.2 & 11.4 & 16.8 & 37.4 & 37.4 & 37.2  \\
       & $Det^2$ & 64.3 & 72.8 & 56.6 & 12.4 & 5.7 & 18.6 & 16.2 & 15.0 & 17.3  & 33.1 & 32.2 & 39.0 \\
       & $Det^3$   & 61.8 & 70.4 & 53.9 & 12.0 & 7.9 & 15.8 & 15.9 & 10.9 & 20.4  & 42.6 & 42.2 & 45.6  \\
         &$Det^4$ & 63.0 & 72.6 & 54.2 & 9.3 & 3.6 & 14.6 & 14.9 & 15.6 & 14.3 & 38.5 & 37.8 & 43.2 \\
        & $Det^5$ & 61.3 & 67.8 & 55.2 & 10.8 & 6.3 & 14.9 & 10.1 & 6.3 & 13.6 & 34.2 & 35.1 & 28.5   \\
       \cmidrule{2-14}
      &$Det^{Mean}_{base}$ 
             & 62.3 & 70.4 & 54.9 & 10.9 & 5.7 & 15.7 & 14.3 & 11.8 & 16.5 &  37.2 & 36.9 & 38.7   \\
     \cmidrule{2-14}
      &$Bayes_{base}$ 
             & 64.0 \textcolor{green}{$\uparrow$}   & 72.9  \textcolor{green}{$\uparrow$}& 55.8 \textcolor{green} {$\uparrow$}  & 10.4 \textcolor{green}{$\downarrow$}  & 5.4 \textcolor{green}{$\downarrow$} & 15.1 \textcolor{green}{$\downarrow$} & 13.1 \textcolor{green}{$\downarrow$} & 10.0 \textcolor{green}{$\downarrow$}  &    16.0 \textcolor{green}{$\downarrow$} & 35.2  \textcolor{green}{$\downarrow$} & 34.7 \textcolor{green}{$\downarrow$} & 38.2 \textcolor{green}{$\downarrow$}   \\
     \hline
      \multirow[]{7}{*} {\begin{tabular}{c}  PET \\$+$ \\ CT \\ $+$ \\ Epi. Uncer\end{tabular}}& $Det^1$ & 64.4 & 72.2 & 57.3 & 13.2 & 8.6 & 17.4 & 8.3 & 5.7 & 10.7 & 43.8 & 43.4 & 46.5  \\
        &$Det^2$ & 64.5 & 71.9 & 57.7 & 14.9 & 10.0 & 19.4 & 8.5 & 6.0 & 10.8 & 45.4 & 45.3 & 46.1\\
       & $Det^3$   & 63.0 & 70.3 & 56.3 & 13.6 & 8.2 & 18.5 & 7.5 & 3.9 & 10.9 & 40.4 & 40.5 & 39.8  \\
        & $Det^4$ & 63.7 & 70.9 & 57.0 & 14.7 & 9.6 & 19.4 & 8.5 & 5.9 & 10.9 & 47.1 & 47.0 & 47.5  \\
        & $Det^5$ & 64.0 & 71.9 & 56.8 & 14.7 & 9.3 & 19.6 & 7.4 & 4.1 & 10.4  & 46.9 & 46.9 & 47.2  \\
       \cmidrule{2-14}
      & $Det^{Mean}_{aug}$ 
             & 63.9 & 71.5 & 57.0 & 14.2 & 9.2 & 18.9 & 8.0 & 5.1 & 10.7 & 44.7 & 44.6 & 45.4  \\
     \cmidrule{2-14}
      & $Bayes_{aug}$ 
             & 64.3 \textcolor{green}{$\uparrow$}   & 71.8  \textcolor{green}{$\uparrow$}& 57.4 \textcolor{green} {$\uparrow$}  & 13.9 \textcolor{green}{$\downarrow$}  & 8.5 \textcolor{green}{$\downarrow$} & 18.8 \textcolor{green}{$\downarrow$} & 8.2 \textcolor{red}{$\uparrow$} & 5.6 \textcolor{red}{$\uparrow$}  &    10.5 \textcolor{green}{$\downarrow$} & 44.8 & 44.7 & 45.8  \\
     \bottomrule
    \end{tabular}
    }
\end{center}
\end{table}

\subsection{Sensitivity analysis of deterministic nnU-Net and Bayesian ensembling}

\begin{figure}[htbp]
\centering
   \includegraphics[width  =1\linewidth]{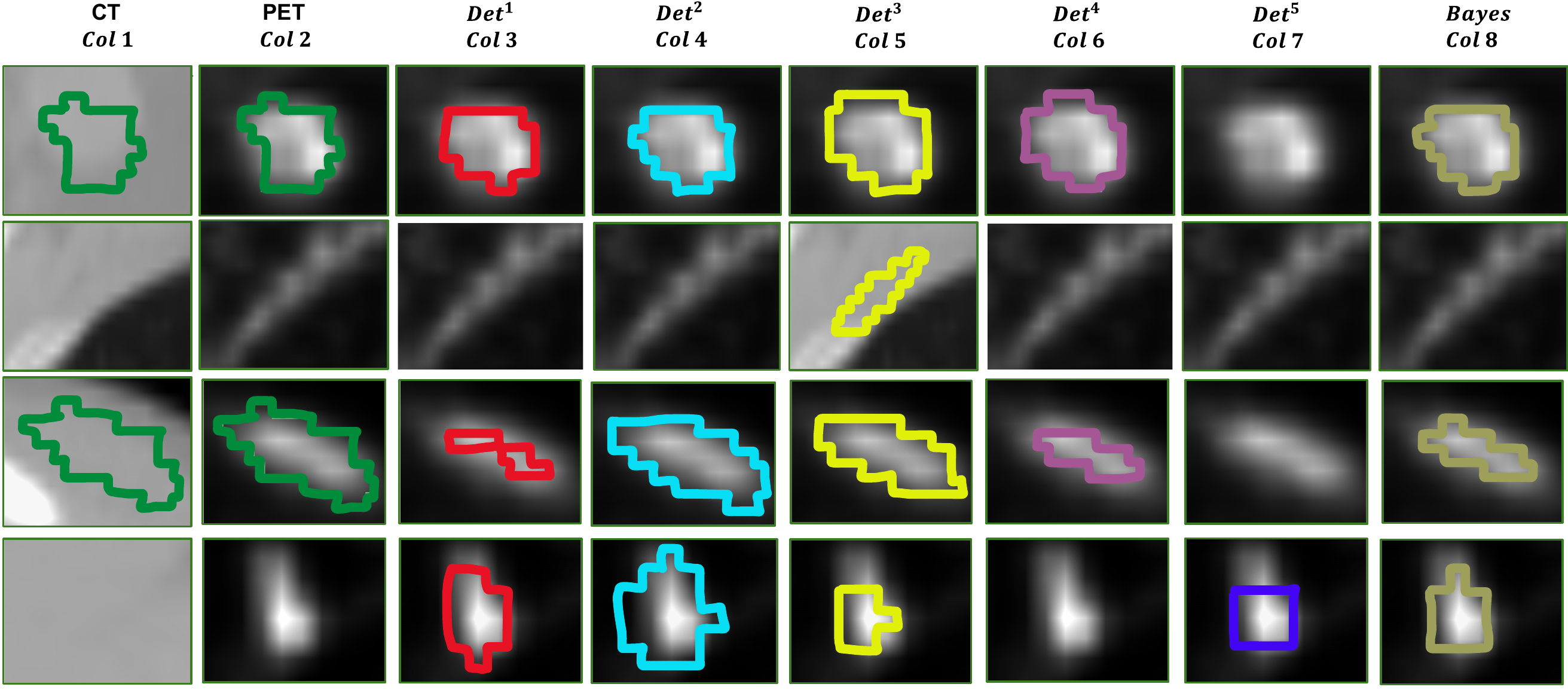}
   \caption[example] 
   { \label{figs:qual_bayes_fdg_nnunet} Deterministic nnU-Net shows variability in predictions, and is sensitive to training stochasticity, whereas Bayesian ensembling improves predictions. 
We visualize four representative regions from two FDG validation cases. Columns 1 and 2 show the CT and PET inputs, and columns 3–7 present the lesion predictions from the five deterministic nnU-Net ensemble members. Column 8 shows the Bayesian ensemble prediction. In the first row, where a lesion is present, four deterministic models ($Det^1$–$Det^4$) correctly identify it as a true positive, and the Bayesian ensemble also produces a true-positive prediction. In the second row, where no lesion is present, $Det^3$ yields a false positive, whereas all other deterministic models and the Bayesian ensemble correctly remain negative. In the third row, $Det^5$ misses the lesion, but the Bayesian ensemble successfully detects it. In the final row, we show a failure case for Bayesian ensembling: four deterministic members produce false positives, and the Bayesian ensemble is unable to correct this error.}
\end{figure}

\textbf{Qualitative evaluation:} Figure \ref{figs:qual_bayes_fdg_nnunet} presents an example with four regions from two FDG PET/CT cases using the nnU-Net \cite{roberts2025prospective} baseline, and we compare deterministic predictions from five deterministic models ($Det^1 \rightarrow Det^5$) with Bayesian ensembling ($Bayes$). Row 1 shows a true-positive case correctly detected by four deterministic models and the Bayesian ensemble, but missed by $Det^5$ model. Row 2 shows a false positive from $Det^3$, while the others remain correctly negative. In Row 3, $Det^5$ misses the lesion, but the Bayesian ensemble detects it. Row 4 presents a failure case of Bayesian ensembling where multiple false positives arise from deterministic members and the Bayesian ensembling is unable to correct the error.

\noindent \textbf{Quantitative evaluation:}  Deterministic models ($Det^1 \rightarrow Det^5$) exhibit substantial variability across DICE, FPVol, and FNVol for both baselines ($Det^1 \rightarrow Det^5$). The Dice scores range from 61.1–64.3 , FPVol ranges from 9.3–12.4, and FNVol ranges from 10.1–16.2. When averaged across deterministic models, the mean deterministic DICE is 62.3 whereas Bayesian model achieves a DICE of 64.0, reducing FPVol from 10.9 → 10.4 and FNVol from 14.3 → 13.1. These results show that deterministic models suffer from substantial training stochasticity, whereas Bayesian ensembling delivers more stable and better segmentation performance. 

\subsection{Uncertainty quantification and its relationship with misclassifications}\label{sec:anal_qual}

\begin{figure} [htbp]
   \begin{center}
   \begin{tabular}{c} 
   \includegraphics[width  =1\linewidth]{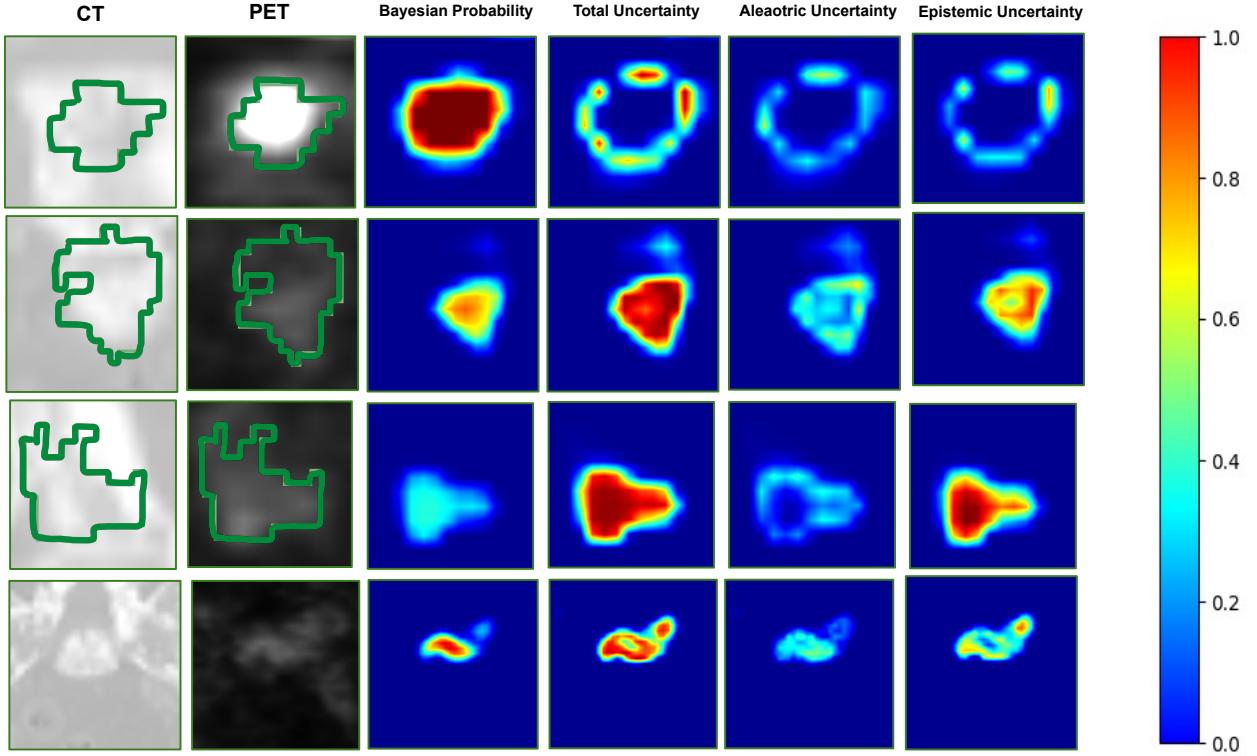}
	\end{tabular}
	\end{center}
   \caption[example] 
   { \label{figs:qual_uncer_fdg_nnUNet} Visualizing Bayesian nnU-Net prediction probabilities and uncertainty maps in four regions of a PSMA sample, highlighting complementary behavior of prediction probability and prediction uncertainty. Row 1 illustrates a true-positive case, where all three uncertainty measures remain low within the lesion core and increase slightly near the boundaries. Row 2 shows a partially captured lesion characterized by uniformly high uncertainty across the predicted region. Row 3 presents a false-negative case in which the Bayesian probability is very low within the ground-truth lesion, yet the uncertainties in that area are elevated. Row 4 depicts a false-positive case exhibiting both high Bayesian probability and high uncertainty throughout the predicted region. It is worth noting that, across all four rows, the aleatoric uncertainty remains substantially lower, indicating that each deterministic model maintains high confidence in its own predictions. }
\end{figure}

\noindent \textbf{Qualitative evaluation}\label{sec:unc_qual}:
Figure \ref{figs:qual_uncer_fdg_nnUNet} shows the total, aleatoric, and epistemic uncertainty maps on a PSMA sample. We observe that for true-positive (TP) lesions, the uncertainty maps primarily highlight the boundary regions, whereas for false-negative (FN) and false-positive (FP) lesions, elevated uncertainty appears throughout the entire lesion region.

\begin{figure}[htbp]
	\centering
	\begin{subfigure}[b]{0.45\textwidth}
	\centering
	\includegraphics[width=\textwidth]{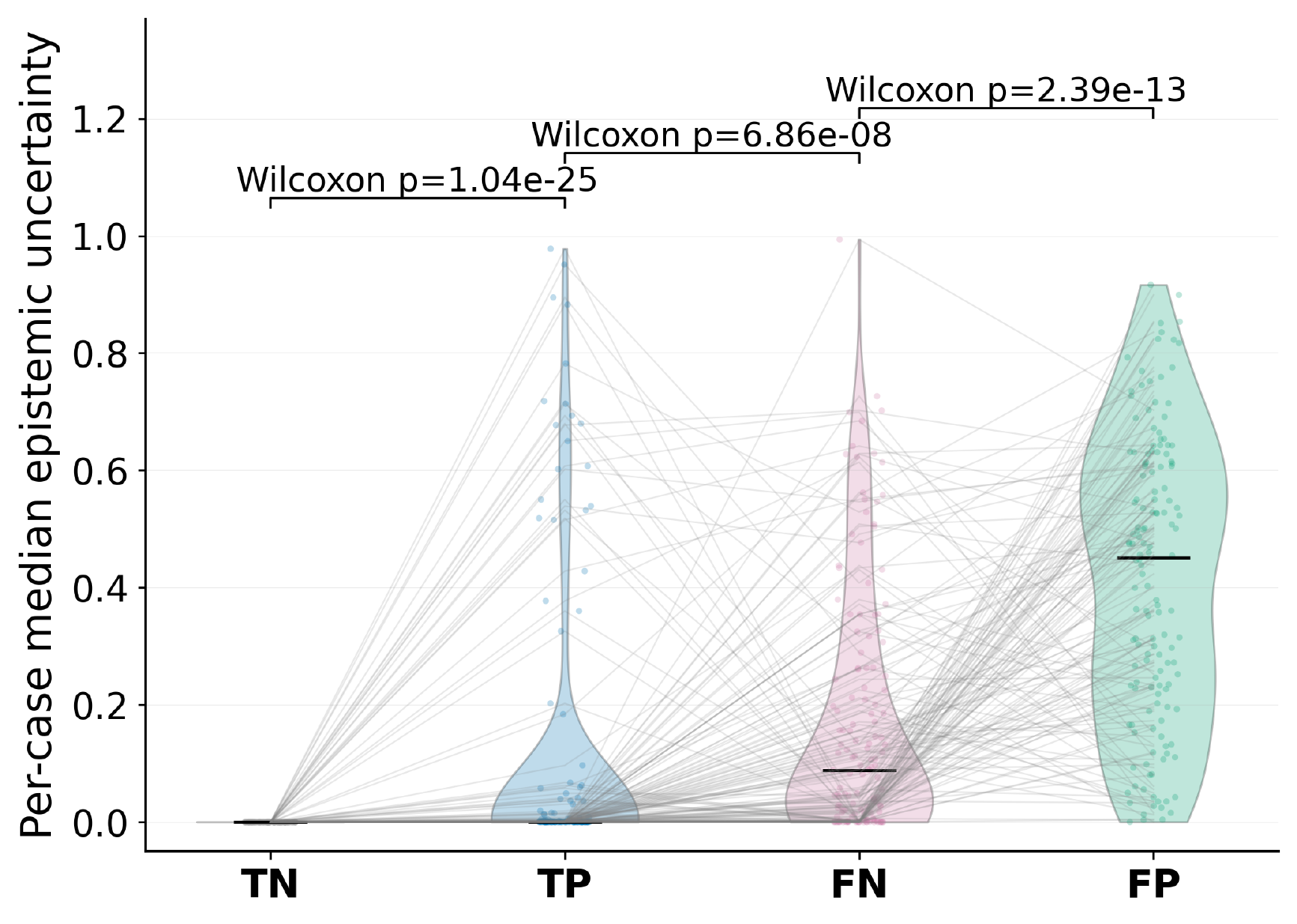}
	\caption{ }
	\label{figs:tn_tp_fn_tp_dist_nnunet}
	\end{subfigure}
	\begin{subfigure}[b]{0.45\textwidth}
		\centering
	\includegraphics[width=\textwidth]{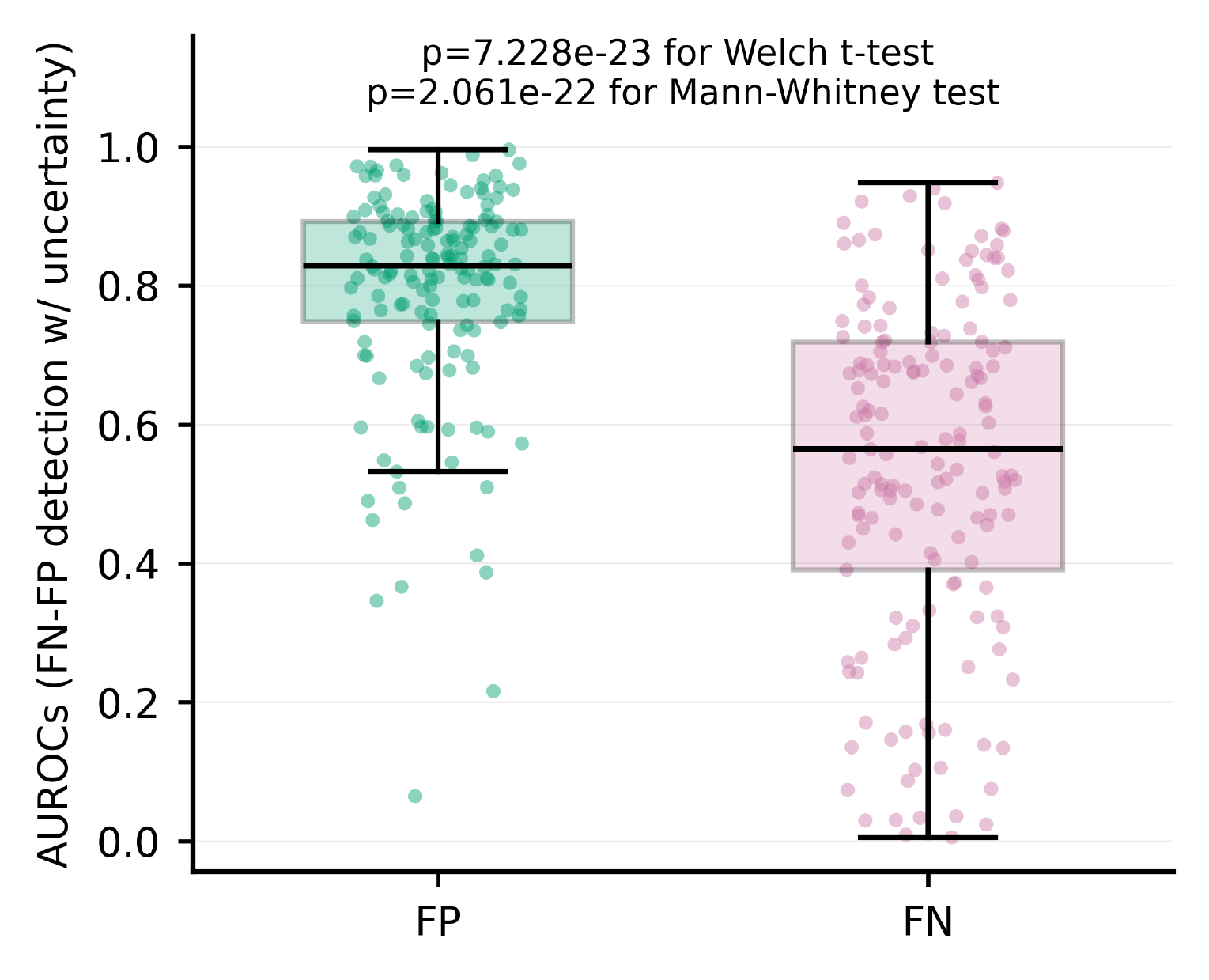}
	\caption{ }
    \label{figs:AUROC_FN_FP_nnunet}
	\end{subfigure}
    \caption[example]{Investigating the relationship between epistemic uncertainty and misclassification shows epistemic uncertainty is significantly high for false positive predictions. (a) Per-case median epistemic uncertainty is higher for misclassified voxels (FP and FN) than for correctly classified voxels (TP and TN). Among the misclassifications, FP voxels exhibit higher per-case median uncertainty than FN voxels. (b) Distributions of AUROC for FP and FN detection with nnUNet \cite{isensee2021nnu}. We treat each misclassified voxel (FP or FN) as $True$ and each correctly classified voxel as $False$, omitting the large number of correctly predicted background voxels (TN). Using epistemic uncertainty as a decision variable, we perform voxel-wise binary classification to determine whether a voxel is misclassified or not. This allows us to compute an ROC curve and the corresponding AUROC for each validation case, providing a measure of how effectively epistemic uncertainty can serve as a thresholding criterion for misclassification detection. AUROC for FP detection shows significantly higher values (p-values $\ll 0.0001$) under both Mann–Whitney and unpaired t-tests. }
    \label{figs:unc_AUROC_AUPRC_nnunet}
\end{figure}

\noindent \textbf{Quantitative evaluation: }\label{sec:unc_err} 
Per-case median epistemic uncertainty is lowest for correctly classified voxels (TN and TP) and significantly higher for misclassified voxels (FN and FP) (Figure \ref{figs:tn_tp_fn_tp_dist_nnunet}). Among the misclassification types, FP voxels exhibit the highest uncertainty, followed by FN, indicating that uncertainty is strongly associated with prediction errors. The statistical tests (Wilcoxon) confirm that the differences between TP–FN and FN–FP distributions are highly significant.

Uncertainty is more effective at identifying false positives than false negatives, as reflected by significantly higher AUROC values for FP detection ($p \ll 0.05$ for both Welch’s t-test and Mann–Whitney test) (Figure \ref{figs:AUROC_FN_FP_nnunet}) . This suggests that uncertainty provides better separability for FP errors, whereas FN errors are comparatively harder to distinguish based on uncertainty alone. Overall, the results demonstrate that epistemic uncertainty is a meaningful indicator of model errors, with stronger discriminative power for FP detection and consistently elevated values in misclassified regions for the baseline nnU-Net model for whole body lesion segmentation.

% \paragraph{AUROC and AUPRC of FN and FP voxels}

% From Figures \ref{figs:AUROC_FN_FP_nnunet} and \ref{figs:AUPRC_FN_FP_nnunet}, we observe a contrasting behaviour of uncertainty in detecting FN and FP voxels. For FP voxels, AUROC values are lower, while AUPRC values are higher for FN voxels. To better understand this seemingly counterintuitive pattern, we analyzed the results in finer detail and observed that, for FN detection, epistemic uncertainty is less reliable for separating FN from TP voxels. Many FN and TP voxels share similarly low uncertainty values, which reduces separability in the low-uncertainty range and leads to lower AUROC scores. However, when uncertainty is high, it is more specifically associated with FN errors, resulting in higher AUPRC values. In other words, uncertainty maps tend to localize FN regions more precisely but fail to capture many FN voxels. In contrast, FP regions are less precise (reflected by lower AUPRC) but exhibit broader, more reliable high-uncertainty regions that yield stronger AUROC performance.

\subsection{Leveraging uncertainty to improve model performance}\label{sec:abla_aug} 

% \subsubsection{Qualitative Evaluation of Uncertainty Augmented Baselines}

% 

\subsubsection{Uncertainty-augmented Bayesian nnU-Net} \label{sec:anal_quant}

\begin{figure} [htbp]
   \begin{center}
   \begin{tabular}{c} 
   \includegraphics[width  =1\linewidth]{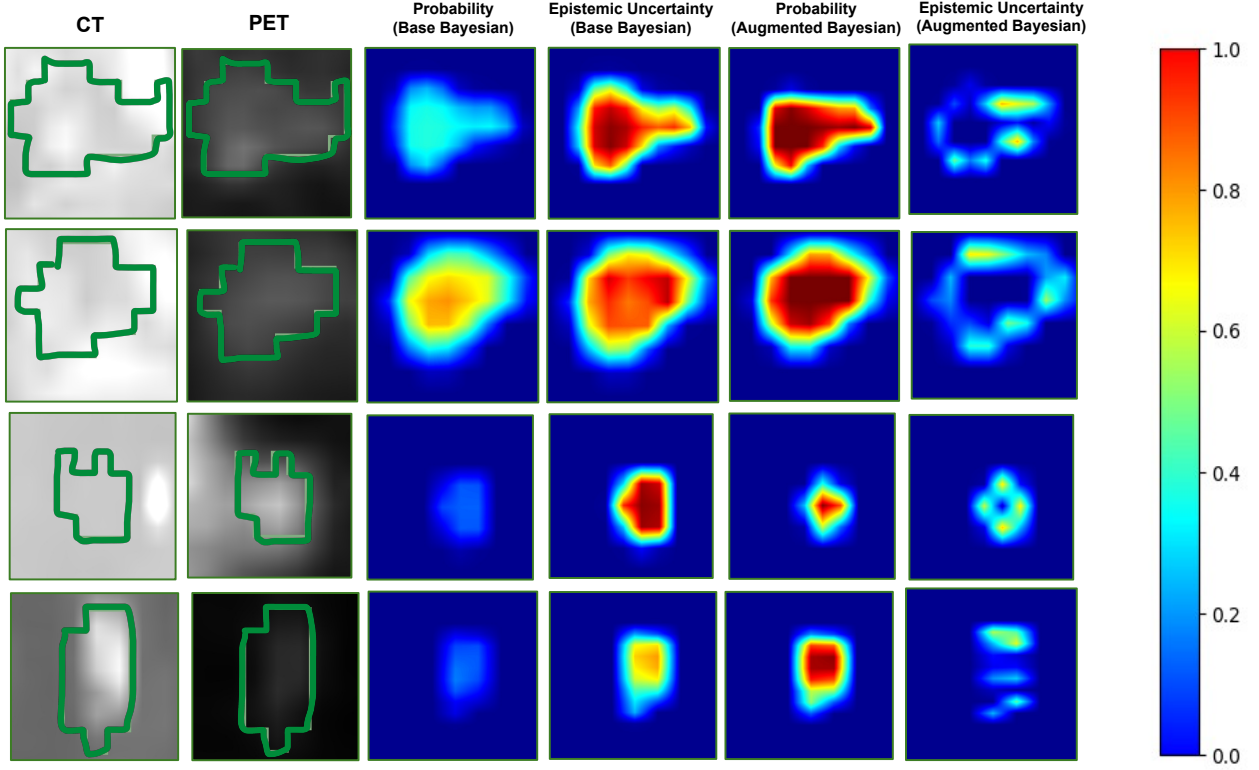}
	\end{tabular}
	\end{center}
   \caption[example]{Uncertainty-augmented Bayesian nnU-Net (Augmented Bayesian) improves lesion detection over baseline Bayesian nnUNet (Base Bayesian) and lowers prediction uncertainty. We visualize four lesions from different PSMA cases to demonstrate how uncertainty augmentation improves performance. In all examples, the base Bayesian model renders low probabilities, leading to false-negative predictions accompanied by high epistemic uncertainty around the lesion regions. With the augmented model, the Bayesian probabilities increase significantly, and the corresponding epistemic uncertainty decreases, indicating more confident and accurate lesion detection.} 
   { \label{figs:qual_uncer_psma} 
}
\end{figure}

\noindent \textbf{Qualitative evaluation:} Figure \ref{figs:qual_uncer_psma} presents four PSMA lesion cases. In all instances, the Bayesian probabilities from the base models (third column) are low, indicating false negatives in the two-channel segmentation model. Notably, the corresponding epistemic uncertainty maps (fourth column) show elevated uncertainty across these lesion regions. In the fifth column, the uncertainty-augmented models successfully recover these missed regions, as reflected by higher Bayesian probabilities. Furthermore, the epistemic uncertainty decreases in the augmented models (last column), indicating improved confidence in the corrected predictions.

\begin{figure} [htbp]
   \begin{center}
   \begin{tabular}{c} 
   \includegraphics[width  =1\linewidth]{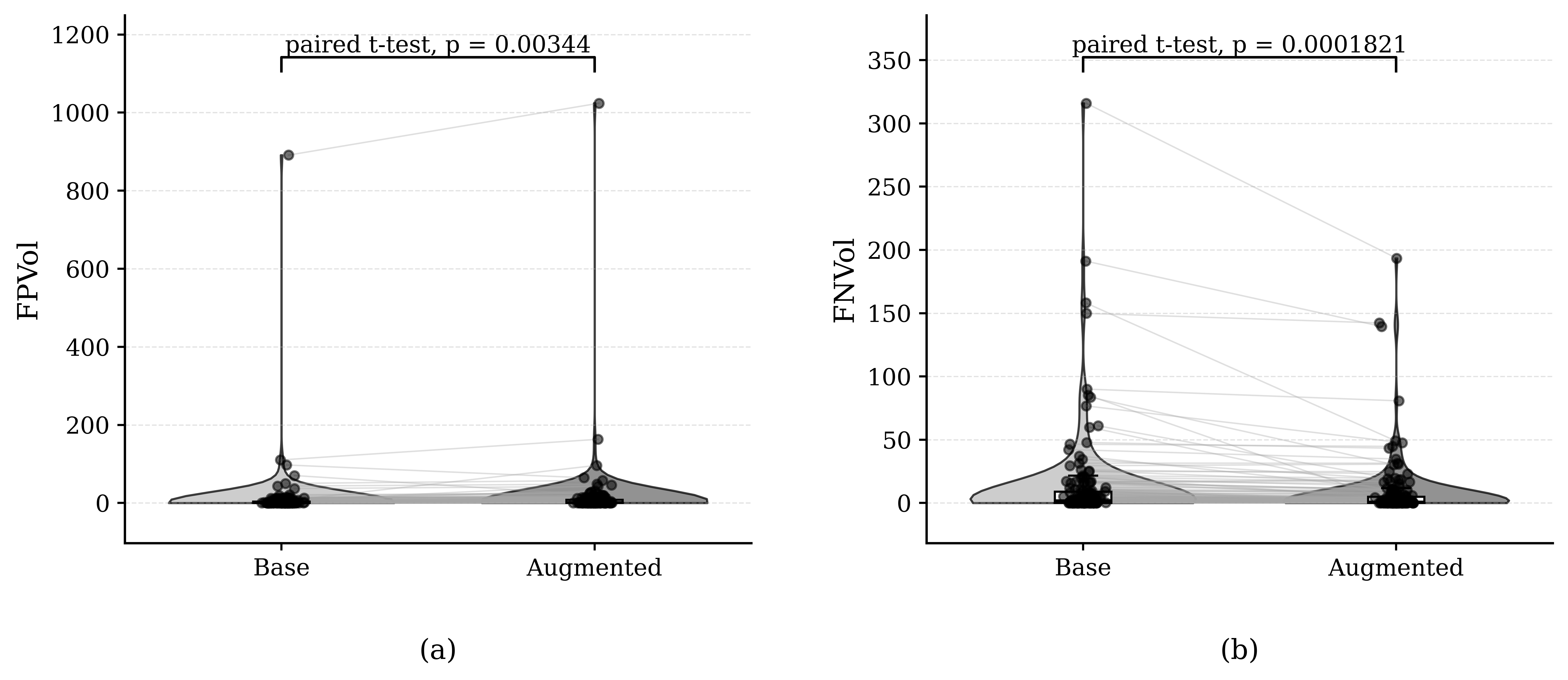}
	\end{tabular}
	\end{center}
   \caption[example]{Complementary behavior between the base (Bayesian) and uncertainty-augmented models across cases. The violin plots illustrate paired distributions of (a) false positive volume (FPVol) and (b) false negative volume (FNVol). The augmented model exhibits a statistically significant increase in FPVol (p = 0.00344), indicating reduced precision, while achieving a significant decrease in FNVol (p = 0.00018), reflecting improved recall. This consistent precision–recall trade-off highlights the complementary strengths of the two models and motivates the need for a case-adaptive routing strategy based on input characteristics (e.g., PET uptake patterns) to selectively leverage each model.} 
   { \label{figs:comple_fpvol_fnvol} 
}
\end{figure}

\noindent \textbf{Quantitative evaluation:} In this subsection, we evaluate whether Bayesian ensembling improves the uncertainty-augmented 3-channel deterministic models, similar to the 2-channel setting in Section \ref{sec:anal_qual}.  Augmented Bayesian model ($Bayes_{aug}$) consistently outperforms the average deterministic counterpart ($Det^{Mean}_{aug}$) (Table \ref{tab:scores_nnunet_bayes}), further supporting \textbf{RQ1} on the benefits of Bayesian ensembling.
Although individual deterministic models show noticeable variability, the mean augmented model ($Det^{Mean}_{aug}$) achieves improved Dice, particularly for PSMA cases (57.0), compared to the base setting. This improvement is driven by a reduction in FNVol (e.g., 10.7 for PSMA with augmented model compared to 16.5 with the base model), indicating better recovery of missed lesions, but it comes with an increase in FPVol (e.g., 18.9 for PSMA with augmented model increasing from 15.7 with base model), reflecting a precision–recall trade-off. The Bayesian ensemble ($Bayes_{aug}$) further stabilizes these predictions and provides additional gains in overall Dice (64.3), while maintaining the same trade-off between FNVol and FPVol. We observe that uncertainty-augmented model achieve significantly lower FNVol compared to base in Figure \ref{figs:comple_fpvol_fnvol}. 

\begin{figure}[htbp]
	\centering

	\begin{minipage}[t]{0.45\textwidth}
		\centering
		\textbf{Base Bayesian}\\[0.5em]

		\begin{subfigure}[b]{\textwidth}
			\centering
			\includegraphics[width=\textwidth]{per_case_tn_tp_fn_fp_uncertainty_violin.pdf}
			\caption{}
			\label{figs:tn_tp_fn_fp_dist_base}
		\end{subfigure}

		\vspace{0.5em}

		\begin{subfigure}[b]{\textwidth}
			\centering
			\includegraphics[width=\textwidth]{boxplot_auroc_all_cases_fp_vs_fn.pdf}
			\caption{}
			\label{figs:auroc_fp_fn_base}
		\end{subfigure}
	\end{minipage}
	\hfill
	\begin{minipage}[t]{0.45\textwidth}
		\centering
		\textbf{Augmented Bayesian}\\[0.5em]

		\begin{subfigure}[b]{\textwidth}
			\centering
			\includegraphics[width=\textwidth]{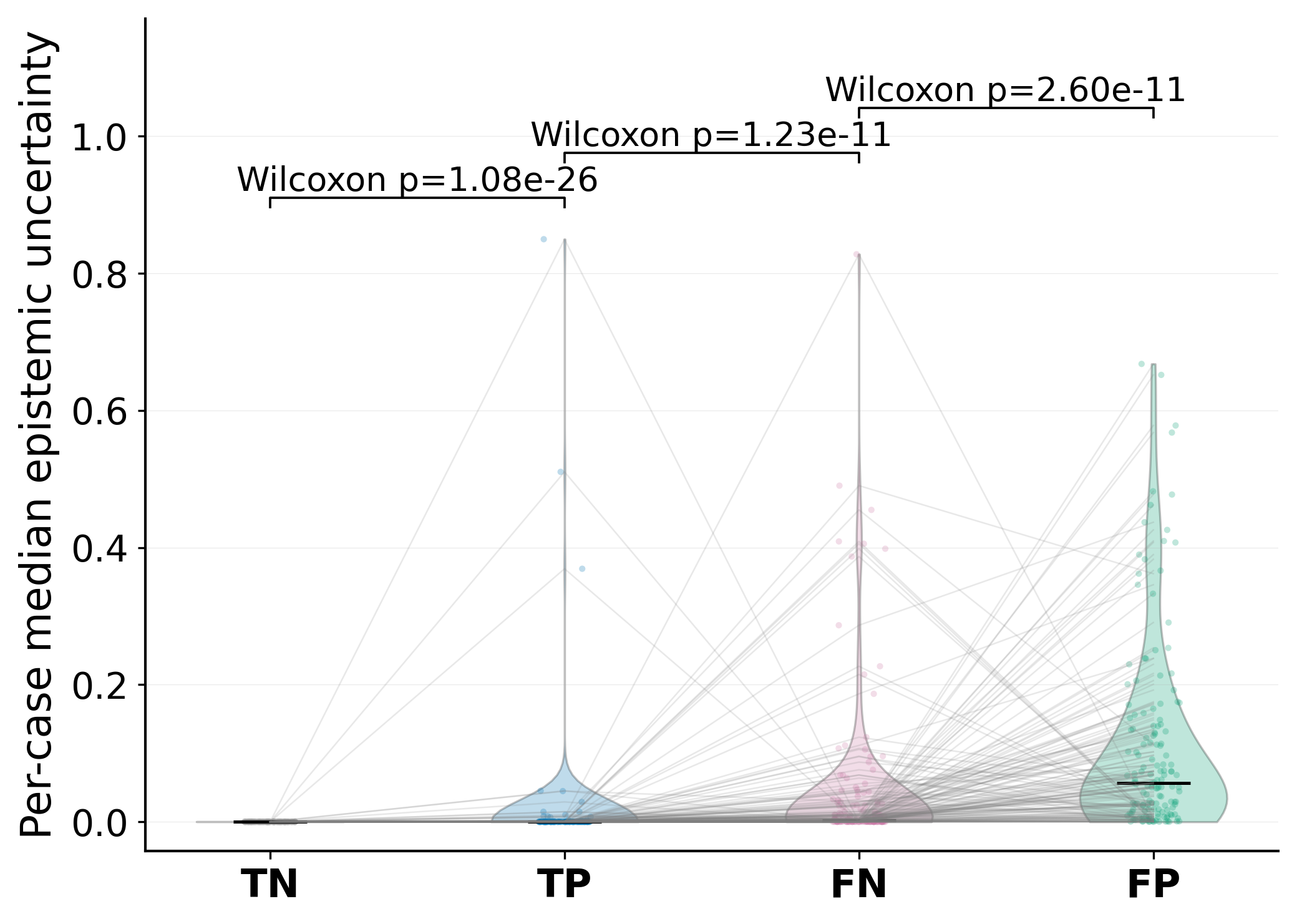}
			\caption{}
			\label{figs:tn_tp_fn_fp_dist_aug}
		\end{subfigure}

		\vspace{0.5em}

		\begin{subfigure}[b]{\textwidth}
			\centering
			\includegraphics[width=\textwidth]{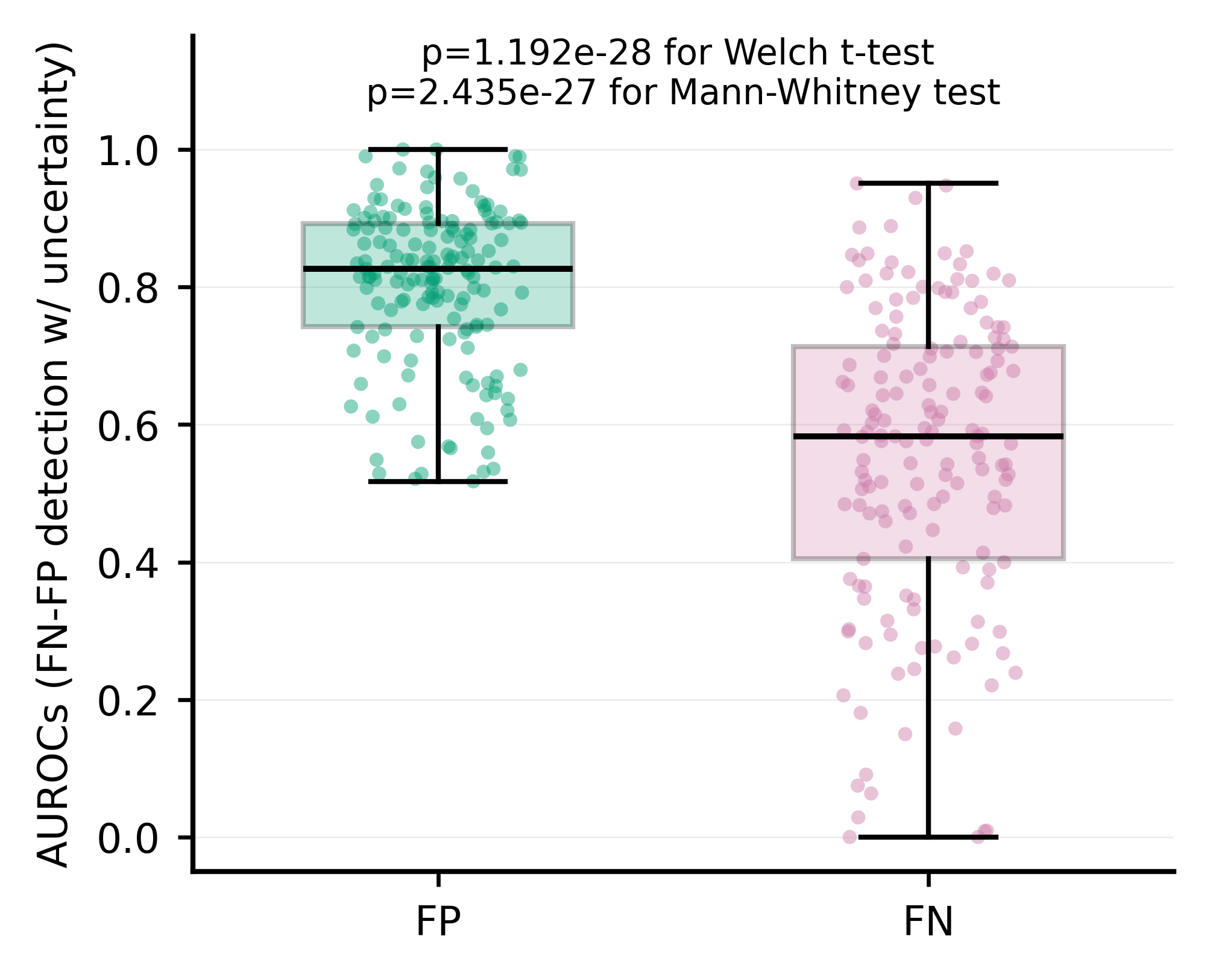}
			\caption{}
			\label{figs:auroc_fp_fn_aug}
		\end{subfigure}
	\end{minipage}

	\caption[example]{Uncertainty--misclassification relationship is consistent between baseline Bayesian (Base Bayesian) and uncertainty-augmented Bayesian (Augmented Bayesian) nnUNet models for whole-body PET/CT lesion segmentation, with overall less uncertainty and fewer outliers in the Augmented Bayesian model. (a,b) Per-case median epistemic uncertainty distributions for true negatives (TN), true positives (TP), false negatives (FN), and false positives (FP) using the (a) base Bayesian model and (b) uncertainty-augmented Bayesian model. Gray lines connect class-specific measurements from the same case, and Wilcoxon test p-values indicate significant differences in uncertainty between misclassified and correctly classified regions. In both models, FP and FN regions exhibit higher uncertainty than TN regions, supporting the association between epistemic uncertainty and segmentation misclassifications. (c,d) AUROC distributions for uncertainty-based detection of FP and FN regions for the (c) base Bayesian model and (d) uncertainty-augmented Bayesian model. FP detection shows consistently higher AUROC than FN detection, with significant differences by Welch's t-test and Mann--Whitney test, indicating that epistemic uncertainty is more discriminative for false-positive errors than for missed lesions.}
	\label{figs:unc_AUROC_AUPRC_deeppsma}
\end{figure}

Similar to the analysis of base model in Sect. \ref{sec:anal_qual}, we also analyze the uncertainty–misclassification relationship for the uncertainty-augmented model. The comparison is presented in Figure \ref{figs:unc_AUROC_AUPRC_deeppsma}. We again observe that uncertainty is more reliable for FP detection than FN detection, as reflected by higher AUROC scores. Interestingly, the AUROC distributions for the uncertainty-augmented model appear less variable than those of the base model for both FP and FN detection.  

\begin{figure} [htbp]
   \begin{center}
   \begin{tabular}{c} 
   \includegraphics[width  =1\linewidth]{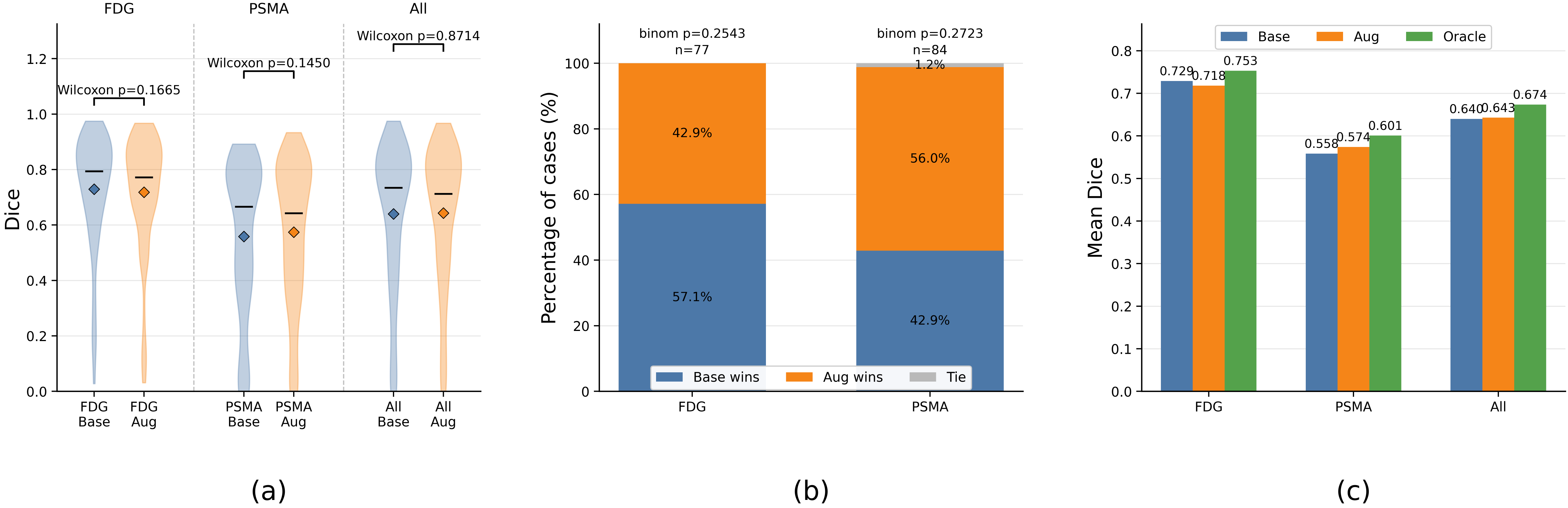}
	\end{tabular}
	\end{center}
   \caption[example]{Complementary performance between the base (Bayesian) and uncertainty-augmented models across tracers. (a) Violin plots of Dice scores show a weak but consistent tracer-dependent trend: for FDG, the base model achieves slightly higher mean Dice than the augmented model, whereas for PSMA, the augmented model slightly outperforms the base model. However, these differences are not statistically significant (Wilcoxon test). (b) Case-wise win-rate analysis further highlights this complementary behavior: the base model performs better in more than $50\%$ of FDG cases, while the augmented model wins in more than $50\%$ of PSMA cases. (c) The oracle performance (selecting the better prediction per case) yields noticeable improvements over both individual models across FDG, PSMA, and all cases, indicating substantial unrealized potential and motivating the need for a case-adaptive routing strategy.} 
   { \label{figs:comple_fdg_psma} 
}
\end{figure}

Moreover, we observe a weak complementary trends between FDG and PSMA in Figure \ref{figs:comple_fdg_psma}. Specifically, FDG samples tend to perform better with the base model, whereas PSMA samples show improved performance with the augmented model. However, this trend is not statistically significant. We further analyze the proportion of win cases and observe that more than $50\%$ of FDG cases favor the base model, while the opposite holds for PSMA.

\subsubsection{Case adaptive routing}\label{sec:abla_carbu}

% \noindent \textbf{Quantitative evaluation:} Table \ref{tab:scores_tambu_nnunet_val} summarizes the fusion and routing results. Among late fusion methods, maximum probability performs best (Dice: 64.8), outperforming intersection and average fusion, suggesting that voxel-wise fusion partially captures the complementary behavior of the base and augmented models. Rule-based tracer-aware routing achieves comparable performance (Dice: 64.8), consistent with the weak, non-significant tracer-dependent differences observed earlier.

\noindent \textbf{Quantitative evaluation:} Table \ref{tab:scores_tambu_nnunet_val} summarizes the fusion and routing results for filtered validation samples. Among late fusion methods, maximum probability performs best (Dice: 61.7), outperforming intersection and average fusion, suggesting that voxel-wise fusion partially captures the complementary behavior of the base and augmented models. Rule-based tracer-aware routing achieves comparable performance (Dice: 61.5), consistent with the weak, non-significant tracer-dependent differences observed earlier.

Learned routing provides the strongest overall performance. The MSE-based router has Dice of 61.6, while CE-based routing performs best, with the $120 \times 120 \times 120$ PET input achieving the highest Dice score of 61.8. This gain reflects improved balancing of false positives and false negatives by adaptively selecting between the precision-oriented base model and recall-oriented augmented model. Across tracers, learned routing with PET as input improves both FDG and PSMA performance, with the largest gains in PSMA Dice, up to 61.8, demonstrating its advantage over heuristic routing and static fusion. 

\begin{table}[htbp]
\centering
\caption {Ablation study on filtered AutoPET-III validation samples ($N=111$) comparing fusion and routing strategies for combining the Bayesian base nnU-Net and uncertainty-augmented nnU-Net models. To reduce noisy supervision, we restrict both training and validation to samples with a Dice difference greater than $0.01$ between the two models. We evaluate three late-fusion strategies and multiple early-routing approaches. The results show that late-fusion methods outperform the individual base and augmented models, while early-routing approaches achieve the best overall performance and further surpass the late-fusion strategies.}
\label{tab:scores_tambu_nnunet_val}
\small
\begin{center}
\resizebox{\textwidth}{!}{
     \begin{tabular}{lllllllllllll}
     \toprule
       
      \multicolumn{2}{c}{\multirow[]{3}{*}{\textbf{Method}}} & 
    \multicolumn{9}{c}{\begin{tabular}{c} Cancer (N=111) \\ FDG=41, PSMA=70
    \end{tabular}}  \\
    \cmidrule(r){3-11} 
    & & \multicolumn{3}{c}{DICE $\uparrow$} & \multicolumn{3}{c}{FPvol $\downarrow$ } & \multicolumn{3}{c}{FNvol $\downarrow$ }  \\
     \cmidrule(r){3-5} \cmidrule(r){6-8} \cmidrule(r){9-11} 
     & & All & FDG & PSMA & All & FDG & PSMA & All & FDG & PSMA  \\
    \midrule
    
     \multirow[]{2}{*}{$Bayes$} &Base (PET + CT) 
             & 60.4 & 66.4 & 56.8 & 13.6 & 7.7 & 17.0 & 12.7 & 8.4 & 15.3  \\
      & Aug (PET + CT + Unc.) & 61.0 & 65.1 & 58.5 & 16.5 & 11.4 & 19.5 & 7.2 & 3.0 & 9.6   \\
   \hline
    % & Uncer. Aug. Baseline  & 62.0  & 68.1 &  \textbf{56.7}    &  10.5 &  9.8&  11.6& 6.3 & 2.9 &    11.8   \\
     %\cmidrule(r){2-11}
      \multirow[]{4}{*}{\textbf{Late Fusion}} & Intersection. & 59.6 & 65.9 & 55.9 & 12.8 & 7.3 & 16.0 & 12.9 & 8.4 & 15.5\\
     & Average Prob. & 61.2 & 65.6 & 58.7 & 16.5 & 12.5 & 18.8 & 7.2 & 3.1 & 9.6 \\
     & Max Prob. & 61.7 & 65.7 & 59.3 & 17.2 & 11.7 & 20.4 & 7.0 & 3.0 & 9.4\\
     \hline
     \multirow[]{9}{*}{\textbf{Early Routing}} & Tracer-aware routing rule &61.5 & 66.4 & 58.5 & 15.1 & 7.7 & 19.5 & 9.2 & 8.4 & 9.6 \\
     & Learnt (CE - 64x64x64) & 61.3 & 66.0 & 58.5 & 15.2 & 8.0 & 19.5 & 7.9 & 5.1 & 9.6\\
     & Learnt (CE - 80x80x80)  & 61.7 & 66.4 & 59.0 & 14.4 & 7.7 & 18.3 & 9.2 & 8.4 & 9.7   \\
     & Learnt (CE - 120x120x120)  & 61.8 & 67.1 & 58.7 & 15.2 & 7.9 & 19.4 & 8.1 & 5.4 & 9.7  \\
     & Learnt (MSE) - PET Only & 61.6 & 66.4 & 58.8 & 14.5 & 8.1 & 18.3 & 9.2 & 8.1 & 9.8\\
     & Learnt (Weighted MSE) - CT Only  & 61.6 & 66.9 & 58.5 & 15.2 & 8.0 & 19.4 & 9.1 & 7.9 & 9.8\\
     & Learnt (Weighted MSE) - PET \& CT & 60.9 & 65.6 & 58.2 & 15.5 & 9.3 & 19.1 & 8.0 & 5.0 & 9.8 \\
     & Learnt (Weighted MSE) - PET \& Uncer. & 60.9 & 65.6 & 58.2 & 15.5 & 9.3 & 19.1 & 8.0 & 5.0 & 9.8\\
     & Learnt (Weighted MSE) - PET Only & \textbf{61.8} & 66.4 & 59.1 & 14.2 & 7.7 & 18.1 & 9.4 & 8.4 & 10.0 \\
  
    % & & \multicolumn{2}{c}{Learnt (CE - 64x64x64)}  & \textbf{65.2} & 72.9 & \textbf{58.2} & 11.4 & 5.4 & 17.0 & 10.3 & 10.0 & 10.5\\
     \bottomrule
    \end{tabular}
    }
\end{center}
\end{table}

% following only if there is an appendix
% \section*{Appendix}
% \addcontentsline{toc}{section}{\numberline{}Appendix}
% Appendix text goes here if needed.

\section{Performance on unseen datasets}

Our ablation experiments and results conducted on the validation set of AutoPet-III in the previous subsection address our first three research questions from \textbf{RQ1} to \textbf{RQ3}. To assess generalizability, we evaluate our methods on two unseen test sets: $321$ held-out AutoPET-III scans for in-distribution testing and $200$ Deep-PSMA scans for OOD testing (Sect. \ref{sec:dataset}).

\begin{figure} [htbp]
   \begin{center}
   \begin{tabular}{c} 
   \includegraphics[width  =1\linewidth]{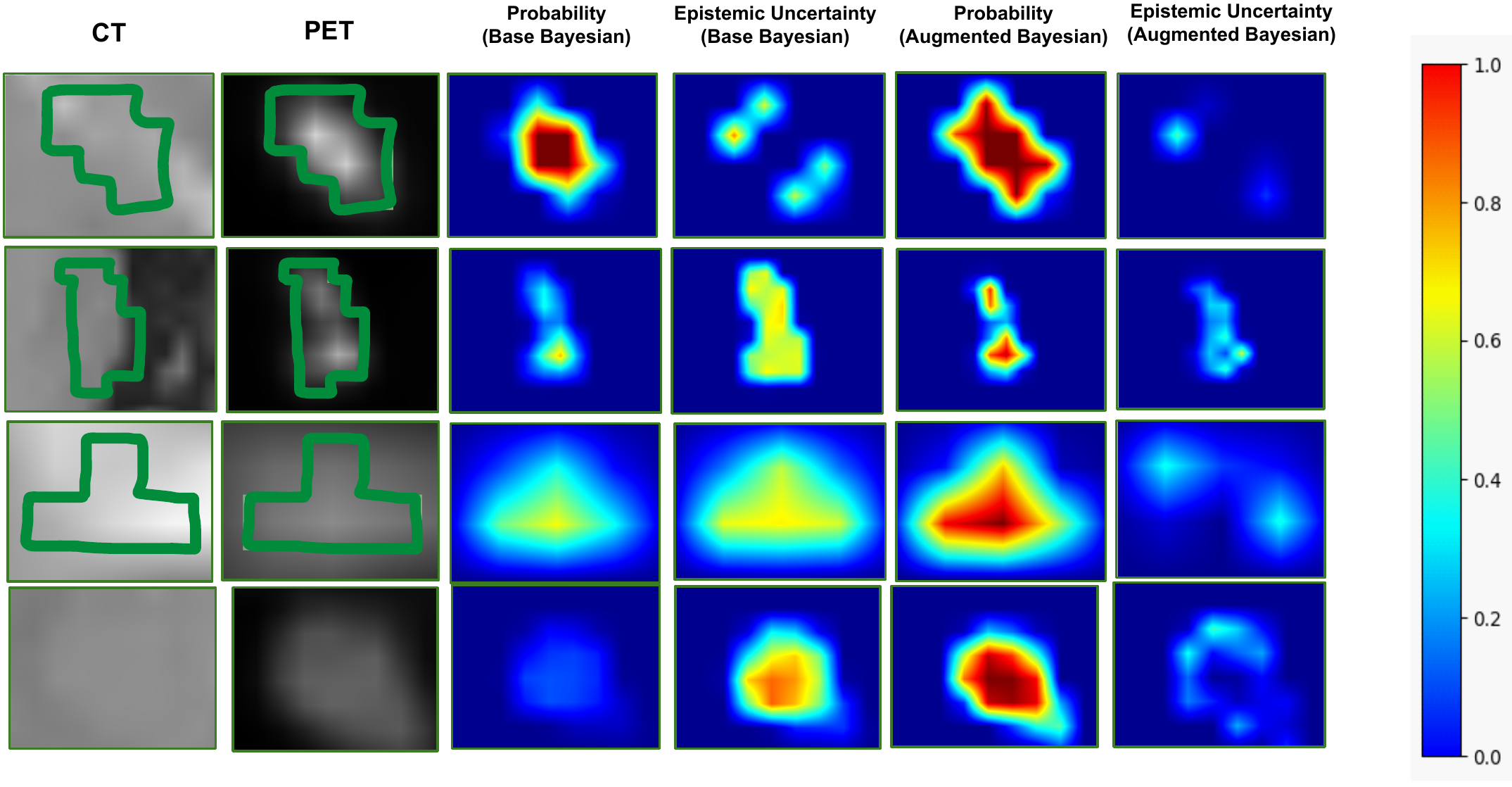}
	\end{tabular}
	\end{center}
   \caption[example]{Uncertainty-augemented Bayesian nnU-net (Augemneted Bayesian) performance generalizes to unseen DeepPSMA dataset. We visualize four lesions from different FDG (first and second row) and PSMA (third and fourth row) cases in the Deep-PSMA dataset to illustrate how uncertainty augmentation improves performance on unseen data. In the first three examples, the base Bayesian model produces low lesion probabilities, resulting in false-negative predictions with elevated epistemic uncertainty around the lesion regions. In contrast, the uncertainty-augmented model substantially increases the Bayesian probabilities while reducing the corresponding epistemic uncertainty, indicating more confident and accurate lesion detection. The last row shows a failure case, where high epistemic uncertainty in a lesion-free region leads the uncertainty-augmented model to generate a false-positive segmentation.} 
   { \label{figs:qual_uncer_deep_psma} 
}
\end{figure}

\noindent \textbf{Qualitative evaluation: } Similar to Figure \ref{figs:qual_uncer_psma}, we qualitatively evaluate the effect of uncertainty augmentation on the Deep-PSMA dataset in Figure Figure \ref{figs:qual_uncer_deep_psma}. The first three rows show successful cases in which the uncertainty-augmented model detects lesions more effectively than the base model. The last row illustrates a failure case, where a high-uncertainty true-negative region in the base model is incorrectly segmented as a false positive by the uncertainty-augmented model.

\noindent \textbf{Quantitative evaluation:} Bayesian ensembling consistently outperforms deterministic models for both the base (PET+CT) and uncertainty-augmented (PET+CT+Unc.) settings, improving Dice and generally reducing FNVol in both the held out AutoPet-III test set, as well as the unseen DeepPSMA dataset (Table \ref{tab:scores_carbu_nnunet_unseen}). On AutoPET-III test set, uncertainty augmentation shows a precision–recall trade-off, reducing FNVol but increasing FPVol. This improves PSMA Dice while slightly lowering FDG performance, consistent with \textbf{RQ2}. The learned router achieves the best overall Dice across ALL, FDG, and PSMA cases, confirming that it effectively exploits the complementary strengths of the base and augmented models, as observed in RQ3.

On Deep-PSMA, uncertainty augmentation improves Dice for both FDG and PSMA cases, with lower FNVol but higher FPVol (Table \ref{tab:scores_carbu_nnunet_unseen}). However, the AutoPET-III-trained router does not outperform the augmented model, suggesting limited OOD transferability of the learned routing strategy.

\begin{table}[htbp]
\centering
\caption {Held-out test performance of the base, uncertainty-augmented, and best-performing case-adaptive routing models on both AutoPET-III ($N=321$) and Deep-PSMA ($N=200$) datasets. In the AutoPET-III test set, there are 207 cancer cases (97 FDG and 110 PSMA) and 114 no-disease cases (102 FDG and 12 PSMA). The Deep-PSMA dataset consists of 200 cancer cases, with 100 FDG and 100 PSMA studies. Consistent with the validation results, uncertainty augmentation generally reduces FNVol in both deterministic and Bayesian settings, while increasing FPVol. On the AutoPET-III test set, FDG Dice decreases whereas PSMA Dice improves under uncertainty augmentation. Our proposed case-adaptive combination through routing achieves the highest Dice scores compared with both deterministic and Bayesian baselines. On the Deep-PSMA dataset, uncertainty augmentation improves performance for both FDG and PSMA cases, while the learned router shows limited generalization.}
\label{tab:scores_carbu_nnunet_unseen}
\small
\begin{center}
\resizebox{\textwidth}{!}{
     \begin{tabular}{lllllllllllllllll}
     \toprule
      \multirow[]{2}{*}{\textbf{Dataset}} & 
      \multicolumn{2}{c}{\multirow[]{2}{*}{\textbf{Method}}}  & \multirow[]{2}{*}{\textbf{Inputs}}&  \multicolumn{3}{c}{DICE $\uparrow$} & \multicolumn{3}{c}{FPvol $\downarrow$ } & \multicolumn{3}{c}{FNvol $\downarrow$ } & \multicolumn{3}{c}{FPvol (No-disease) $\downarrow$ }  \\
     \cmidrule(r){5-7} \cmidrule(r){8-10} \cmidrule(r){11-13}\cmidrule(r){14-16}
   & & & & All & FDG & PSMA & All & FDG & PSMA & All & FDG & PSMA & All & FDG & PSMA  \\
    \midrule
    \multirow[]{5}{*}{AutoPET-III} 
    &  \multirow[]{2}{*}{$Det^{Mean}$} &  Base & PET + CT&  60.1 & 68.2 & 52.9 & 2.9 & 4.2 & 1.9 & 13.9 & 10.9 & 16.6 & 14.4 & 8.4 & 65.4\\
           & & Aug & PET + CT + Unc. & 61.5 & 67.5 & 56.1 & 4.9 \textcolor{red} {$\uparrow$} & 6.9 \textcolor{red} {$\uparrow$} & 3.1 \textcolor{red} {$\uparrow$} & 9.5 \textcolor{green} {$\downarrow$}  & 5.7 \textcolor{green} {$\downarrow$} & 12.8 \textcolor{green} {$\downarrow$} & 21.1 \textcolor{red} {$\uparrow$} & 13.7 \textcolor{red} {$\uparrow$} & 84.2 \textcolor{red} {$\uparrow$} \\
           \cmidrule(r){2-16}
    & \multirow[]{2}{*}{$Bayes$} &Base & PET + CT 
             & 62.0 & 70.7 & 54.3 & 2.0 & 2.9 & 1.3 & 13.6 & 10.3 & 16.4 &12.3 & 6.4 & 62.0  \\
     & & Aug & PET + CT + Unc. & 61.7 & 67.8 & 56.2 & 4.8 \textcolor{red} {$\uparrow$} & 6.8 \textcolor{red} {$\uparrow$} & 3.0 \textcolor{red} {$\uparrow$} & 9.3 \textcolor{green} {$\downarrow$} & 5.6  \textcolor{green} {$\downarrow$} & 12.6   \textcolor{green} {$\downarrow$} &21.1 \textcolor{red} {$\uparrow$} & 13.5 \textcolor{red} {$\uparrow$} & 85.1 \textcolor{red} {$\uparrow$}   \\
    \cmidrule(r){2-16}
    % & Uncer. Aug. Baseline  & 62.0  & 68.1 &  \textbf{56.7}    &  10.5 &  9.8&  11.6& 6.3 & 2.9 &    11.8   \\
     %\cmidrule(r){2-11}
    & \textbf{Combination} & \multicolumn{2}{c}{Learnt router (Weighted MSE; PET)} & \textbf{63.1} & \textbf{70.7} & \textbf{56.3} & 2.9 & 2.9 & 2.9 & 11.6 & 10.3 & 12.8 & 14.7   &  6.4 &   85.1 \\
    \midrule
    \multirow[]{5}{*}{Deep PSMA}
           & \multirow[]{2}{*}{$Det^{Mean}$} &  Base & PET + CT&  62.8 & 60.9 & 64.6 & 8.3 & 4.8 & 11.7 & 19.3 & 23.1 & 15.6& -- & -- & --\\
           & & Aug & PET + CT + Unc. & 65.1  & 63.7   & 66.4   & 17.5 \textcolor{red} {$\uparrow$} & 7.4 \textcolor{red} {$\uparrow$} & 27.6 \textcolor{red} {$\uparrow$} & 11.3 \textcolor{green} {$\downarrow$} & 13.2 \textcolor{green} {$\downarrow$}& 9.4 \textcolor{green} {$\downarrow$} & -- & -- & --  \\
           \cmidrule(r){2-13}
    & \multirow[]{2}{*}{$Bayes$} &Base & PET + CT 
             & 64.7 & 63.8 & 65.5 & 5.6 & 3.4 & 7.8 & 17.9 & 22.3 & 13.5 & -- & -- & --  \\
     & & Aug & PET + CT + Unc. & \textbf{65.3}  & \textbf{63.9}  & \textbf{66.8}  & 17.9 \textcolor{red} {$\uparrow$}  & 7.4 \textcolor{red} {$\uparrow$}  & 28.3 \textcolor{red} {$\uparrow$}  & 11.4 \textcolor{green} {$\downarrow$} & 13.5 \textcolor{green} {$\downarrow$} & 9.2 \textcolor{green} {$\downarrow$}  & -- & -- & --  \\
    \cmidrule(r){2-13}
    & \textbf{Combination} & \multicolumn{2}{c}{Learnt router {Weighted MSE; PET}}  & 65.1 & 63.9 & 66.2 & 14.0 & 3.7 & 24.3 & 16.2 & 22.1 & 10.4 & -- & -- & --\\
     \bottomrule
    \end{tabular}
    }
\end{center}
\end{table}

\subsection{Statistical analysis on the unseen test set}

We conduct statistical comparisons between the best-performing deterministic nnU-Net and our proposed method on the unseen AutoPET-III test set, yielding p-values of 0.022 (t-test) and 0.043 (Wilcoxon signed-rank test). We perform a similar analysis on the Deep-PSMA dataset, comparing the best deterministic baseline with the uncertainty-augmented Bayesian model, where we obtain p-values of 0.00001 (t-test) and 0.00000 (Wilcoxon test), indicating highly significant improvements. 

\section{Discussion}

In this work, we propose a uncertainty-aware framework for multi-tracer, pan-cancer, whole-body PET/CT lesion segmentation. Our approach mitigates the stochastic variability of deterministic nnU-Net models, enables voxel-wise uncertainty quantification, uses uncertainty to recover missed lesions, and improves segmentation performance. Existing deterministic methods are sensitive to random initialization and provide no predictive uncertainty, while heterogeneous multi-tracer datasets can bias models toward low-lesion cases, leading to missed detections in high-burden patients. We therefore investigate four questions: the benefits of Bayesian ensembling, the relationship between uncertainty and misclassification, the value of uncertainty-augmented training, and generalization to in-distribution and OOD test sets.

Our experiments yield four main findings. First, Bayesian ensembling produces more stable and accurate segmentations than individual deterministic models. Second, epistemic uncertainty is more strongly associated with false positives and missed lesions than with correct predictions in nn-UNet models. Third, uncertainty-augmented training improves recall by recovering missed lesions, introducing a precision–recall trade-off with the base model. A learned case-adaptive router further improves performance by selecting the most suitable model for each case. Finally, the uncertainty-augmented model generalizes well to OOD data, improving performance for both FDG and PSMA cases with high lesion burden.

This study has some limitations. First, deep ensembling is computationally expensive, so our experiments are limited to five ensemble members; future work will explore larger ensembles and lightweight uncertainty estimation methods. Second, our analysis is based on nnU-Net, and future studies will extend the framework to other architectures, including other CNN models, vision transformers, diffusion models, and Mamba-based models. Third, our learned router does not generalize well to the Deep-PSMA dataset. Since the router is trained to select the better-performing model based on AutoPET-III-specific input features, the learned decision boundary may not transfer reliably to an external dataset. This effect may be further amplified by the relatively shallow 3D CNN architecture of the router, which may overfit to dataset-specific cues rather than learning domain-invariant routing features. In our future works, we will explore domain-invariant learning of the case-adaptive router. Finally, we only included publicly available datasets and did not include private heterogeneous datasets. Our future work will focus on including private datasets. 

To our knowledge, this is the first study to connect Bayesian deep learning with multi-tracer, pan-cancer, whole-body PET/CT lesion segmentation. We also introduce uncertainty-augmented nnU-Net training and a learnable case-adaptive routing mechanism for selecting between complementary models. By evaluating these components on both in-distribution and OOD test sets, our work provides one of the first systematic assessments of robustness, uncertainty, and generalization in whole-body lesion segmentation.

Our study supports clinical deployment of automated whole-body lesion segmentation by improving robustness with Bayesian ensembling and systematically evaluating the sensitivity of deterministic models. In a multi-tracer, pan-cancer setting, we provide new insights into uncertainty quantification and show that uncertainty-augmented nnU-Net improves detection in high-burden cases, such as Deep-PSMA, with a simple architectural extension. Finally, evaluation on unseen datasets offers an early systematic assessment of generalization in whole-body lesion segmentation.

\section{Conclusion}
In this paper, we present an uncertainty-aware framework for PET/CT-based whole-body lesion segmentation. Our approach begins with Bayesian ensembling of deterministic baselines, yielding improved prediction robustness along with reliable uncertainty quantification. We then analyze the relationship between uncertainty maps and model misclassifications, and leverage these insights to design an uncertainty-augmented training strategy that improves segmentation performance by balancing precision and recall. Furthermore, we introduce a case-adaptive routing mechanism that effectively combines the complementary strengths of the base Bayesian model and the uncertainty-augmented Bayesian model, resulting in a more reliable and generalizable segmentation pipeline. Finally, evaluations on both in-distribution and out-of-distribution unseen test datasets further demonstrate the generalization capability of the proposed framework. Our study enhances the clinical utility of AI-assisted whole-body lesion segmentation by improving robustness, enabling uncertainty-aware modeling through uncertainty quantification, and demonstrating strong generalizability, which broadens its applicability in multi-tracer, pan-cancer clinical workflows, including diagnosis and treatment planning.  

\section*{Acknowledgement Statement}
Research reported in this publication was supported by an Institutional Development Award (IDeA) from the National Institute of General Medical Sciences of the National Institutes of Health under grant number 1P30GM149408. We also gratefully acknowledge support from the Munck-Pfefferkorn Fund, the American Cancer Society Institutional Research Grant, and the Department of Biomedical Data Science at Dartmouth College. All of this support was instrumental in making this work possible.

\bibliographystyle{./medphy.bst}
\bibliography{example}  %%% Uncomment this line and comment out the ``thebibliography'' section below to use the external .bib file (using bibtex) .

@article{oreiller2022head,
  title={Head and neck tumor segmentation in PET/CT: the HECKTOR challenge},
  author={Oreiller, Valentin and Andrearczyk, Vincent and Jreige, Mario and Boughdad, Sarah and Elhalawani, Hesham and Castelli, Joel and Vallieres, Martin and Zhu, Simeng and Xie, Juanying and Peng, Ying and others},
  journal={Medical image analysis},
  volume={77},
  pages={102336},
  year={2022},
  publisher={Elsevier}
}

@article{bagci2013joint,
  title={Joint segmentation of anatomical and functional images: Applications in quantification of lesions from PET, PET-CT, MRI-PET, and MRI-PET-CT images},
  author={Bagci, Ulas and Udupa, Jayaram K and Mendhiratta, Neil and Foster, Brent and Xu, Ziyue and Yao, Jianhua and Chen, Xinjian and Mollura, Daniel J},
  journal={Medical image analysis},
  volume={17},
  number={8},
  pages={929--945},
  year={2013},
  publisher={Elsevier}
}

@article{song2013optimal,
  title={Optimal co-segmentation of tumor in PET-CT images with context information},
  author={Song, Qi and Bai, Junjie and Han, Dongfeng and Bhatia, Sudershan and Sun, Wenqing and Rockey, William and Bayouth, John E and Buatti, John M and Wu, Xiaodong},
  journal={IEEE transactions on medical imaging},
  volume={32},
  number={9},
  pages={1685--1697},
  year={2013},
  publisher={IEEE}
}

@article{yang2015multimodality,
  title={A multimodality segmentation framework for automatic target delineation in head and neck radiotherapy},
  author={Yang, Jinzhong and Beadle, Beth M and Garden, Adam S and Schwartz, David L and Aristophanous, Michalis},
  journal={Medical physics},
  volume={42},
  number={9},
  pages={5310--5320},
  year={2015},
  publisher={Wiley Online Library}
}

@article{yu2008coregistered,
  title={Coregistered FDG PET/CT-based textural characterization of head and neck cancer for radiation treatment planning},
  author={Yu, Huan and Caldwell, Curtis and Mah, Katherine and Mozeg, Daniel},
  journal={IEEE transactions on medical imaging},
  volume={28},
  number={3},
  pages={374--383},
  year={2008},
  publisher={IEEE}
}

@article{jemaa2020tumor,
  title={Tumor segmentation and feature extraction from whole-body FDG-PET/CT using cascaded 2D and 3D convolutional neural networks},
  author={Jemaa, Skander and Fredrickson, Jill and Carano, Richard AD and Nielsen, Tina and de Crespigny, Alex and Bengtsson, Thomas},
  journal={Journal of digital imaging},
  volume={33},
  number={4},
  pages={888--894},
  year={2020},
  publisher={Springer}
}

@inproceedings{milletari2016v,
  title={V-net: Fully convolutional neural networks for volumetric medical image segmentation},
  author={Milletari, Fausto and Navab, Nassir and Ahmadi, Seyed-Ahmad},
  booktitle={2016 fourth international conference on 3D vision (3DV)},
  pages={565--571},
  year={2016},
  organization={Ieee}
}

@article{xu2018automated,
  title={Automated whole-body bone lesion detection for multiple myeloma on 68Ga-pentixafor PET/CT imaging using deep learning methods},
  author={Xu, Lina and Tetteh, Giles and Lipkova, Jana and Zhao, Yu and Li, Hongwei and Christ, Patrick and Piraud, Marie and Buck, Andreas and Shi, Kuangyu and Menze, Bjoern H},
  journal={Contrast media \& molecular imaging},
  volume={2018},
  number={1},
  pages={2391925},
  year={2018},
  publisher={Wiley Online Library}
}

@article{zhao2018tumor,
  title={Tumor co-segmentation in PET/CT using multi-modality fully convolutional neural network},
  author={Zhao, Xiangming and Li, Laquan and Lu, Wei and Tan, Shan},
  journal={Physics in Medicine \& Biology},
  volume={64},
  number={1},
  pages={015011},
  year={2018},
  publisher={IOP Publishing}
}

@inproceedings{jin2019accurate,
  title={Accurate esophageal gross tumor volume segmentation in PET/CT using two-stream chained 3D deep network fusion},
  author={Jin, Dakai and Guo, Dazhou and Ho, Tsung-Ying and Harrison, Adam P and Xiao, Jing and Tseng, Chen-Kan and Lu, Le},
  booktitle={International Conference on Medical Image Computing and Computer-Assisted Intervention},
  pages={182--191},
  year={2019},
  organization={Springer}
}

@article{huang2018fully,
  title={Fully automated delineation of gross tumor volume for head and neck cancer on PET-CT using deep learning: A dual-center study},
  author={Huang, Bin and Chen, Zhewei and Wu, Po-Man and Ye, Yufeng and Feng, Shi-Ting and Wong, Ching-Yee Oliver and Zheng, Liyun and Liu, Yong and Wang, Tianfu and Li, Qiaoliang and others},
  journal={Contrast media \& molecular imaging},
  volume={2018},
  number={1},
  pages={8923028},
  year={2018},
  publisher={Wiley Online Library}
}

@article{blanc2021fully,
  title={Fully automatic segmentation of diffuse large B cell lymphoma lesions on 3D FDG-PET/CT for total metabolic tumour volume prediction using a convolutional neural network.},
  author={Blanc-Durand, Paul and J{\'e}gou, Simon and Kanoun, Salim and Berriolo-Riedinger, Alina and Bodet-Milin, Caroline and Kraeber-Bod{\'e}r{\'e}, Fran{\c{c}}oise and Carlier, Thomas and Le Gouill, Steven and Casasnovas, Ren{\'e}-Olivier and Meignan, Michel and others},
  journal={European Journal of Nuclear Medicine and Molecular Imaging},
  volume={48},
  number={5},
  pages={1362--1370},
  year={2021},
  publisher={Springer}
}

@inproceedings{liu2024lm,
  title={LM-UNet: Whole-Body PET-CT Lesion Segmentation with Dual-Modality-Based Annotations Driven by Latent Mamba U-Net},
  author={Liu, Anglin and Jia, Dengqiang and Sun, Kaicong and Meng, Runqi and Zhao, Meixin and Jiang, Yongluo and Dong, Zhijian and Gao, Yaozong and Shen, Dinggang},
  booktitle={International Conference on Medical Image Computing and Computer-Assisted Intervention},
  pages={405--414},
  year={2024},
  organization={Springer}
}

@inproceedings{song2024dueu,
  title={DuEU-Net: Dual Encoder UNet with Modality-Agnostic Training for PET-CT Multi-modal Organ and Lesion Segmentation},
  author={Song, Jinhong and Yang, Xiao and Liang, Xinglong and Huang, Jiaju and Ma, Junqiang and Sun, Yue and Luo, Wuman and Mok, SengPeng and Wang, Ying and Tan, Tao},
  booktitle={Deep Breast Workshop on AI and Imaging for Diagnostic and Treatment Challenges in Breast Care},
  pages={23--31},
  year={2024},
  organization={Springer}
}

@article{rokuss2024fdg,
  title={From FDG to PSMA: A Hitchhiker's Guide to Multitracer, Multicenter Lesion Segmentation in PET/CT Imaging},
  author={Rokuss, Maximilian and Kovacs, Balint and Kirchhoff, Yannick and Xiao, Shuhan and Ulrich, Constantin and Maier-Hein, Klaus H and Isensee, Fabian},
  journal={arXiv preprint arXiv:2409.09478},
  year={2024}
}

@article{isensee2021nnu,
  title={nnU-Net: a self-configuring method for deep learning-based biomedical image segmentation},
  author={Isensee, Fabian and Jaeger, Paul F and Kohl, Simon AA and Petersen, Jens and Maier-Hein, Klaus H},
  journal={Nature methods},
  volume={18},
  number={2},
  pages={203--211},
  year={2021},
  publisher={Nature Publishing Group}
}

@article{wang2019aleatoric,
  title={Aleatoric uncertainty estimation with test-time augmentation for medical image segmentation with convolutional neural networks},
  author={Wang, Guotai and Li, Wenqi and Aertsen, Michael and Deprest, Jan and Ourselin, S{\'e}bastien and Vercauteren, Tom},
  journal={Neurocomputing},
  volume={338},
  pages={34--45},
  year={2019},
  publisher={Elsevier}
}

@inproceedings{wen2024denoising,
  title={From denoising training to test-time adaptation: Enhancing domain generalization for medical image segmentation},
  author={Wen, Ruxue and Yuan, Hangjie and Ni, Dong and Xiao, Wenbo and Wu, Yaoyao},
  booktitle={Proceedings of the IEEE/CVF Winter Conference on Applications of Computer Vision},
  pages={464--474},
  year={2024}
}

@inproceedings{xu2022improved,
  title={Improved domain generalization for cell detection in histopathology images via test-time stain augmentation},
  author={Xu, Chundan and Wen, Ziqi and Liu, Zhiwen and Ye, Chuyang},
  booktitle={International Conference on Medical Image Computing and Computer-Assisted Intervention},
  pages={150--159},
  year={2022},
  organization={Springer}
}

@inproceedings{liu2020shape,
  title={Shape-aware meta-learning for generalizing prostate MRI segmentation to unseen domains},
  author={Liu, Quande and Dou, Qi and Heng, Pheng-Ann},
  booktitle={International conference on medical image computing and computer-assisted intervention},
  pages={475--485},
  year={2020},
  organization={Springer}
}

@inproceedings{chen2022maxstyle,
  title={Maxstyle: Adversarial style composition for robust medical image segmentation},
  author={Chen, Chen and Li, Zeju and Ouyang, Cheng and Sinclair, Matthew and Bai, Wenjia and Rueckert, Daniel},
  booktitle={International Conference on Medical Image Computing and Computer-Assisted Intervention},
  pages={151--161},
  year={2022},
  organization={Springer}
}

@article{yoon2024domain,
  title={Domain generalization for medical image analysis: A review},
  author={Yoon, Jee Seok and Oh, Kwanseok and Shin, Yooseung and Mazurowski, Maciej A and Suk, Heung-Il},
  journal={Proceedings of the IEEE},
  volume={112},
  number={10},
  pages={1583--1609},
  year={2024},
  publisher={IEEE}
}

@dataset{jeblick2024psmapetctlesions,
  author    = {Jeblick, K. and others},
  title     = {A Whole-Body PSMA-PET/CT Dataset with Manually Annotated Tumor Lesions (PSMA-PET-CT-Lesions)},
  year      = {2024},
  version   = {1},
  publisher = {The Cancer Imaging Archive},
  doi       = {10.7937/r7ep-3x37}
}

@article{gatidis2022whole,
  title={A whole-body FDG-PET/CT dataset with manually annotated tumor lesions},
  author={Gatidis, Sergios and Hepp, Tobias and Fr{\"u}h, Marcel and La Foug{\`e}re, Christian and Nikolaou, Konstantin and Pfannenberg, Christina and Sch{\"o}lkopf, Bernhard and K{\"u}stner, Thomas and Cyran, Clemens and Rubin, Daniel},
  journal={Scientific Data},
  volume={9},
  number={1},
  pages={601},
  year={2022},
  publisher={Nature Publishing Group UK London}
}

@inproceedings{jungo2019assessing,
  title={Assessing reliability and challenges of uncertainty estimations for medical image segmentation},
  author={Jungo, Alain and Reyes, Mauricio},
  booktitle={International Conference on Medical Image Computing and Computer-Assisted Intervention},
  pages={48--56},
  year={2019},
  organization={Springer}
}

@article{yu2024heterogeneity,
  title={Heterogeneity and predictors of the effects of AI assistance on radiologists},
  author={Yu, Feiyang and Moehring, Alex and Banerjee, Oishi and Salz, Tobias and Agarwal, Nikhil and Rajpurkar, Pranav},
  journal={Nature Medicine},
  volume={30},
  number={3},
  pages={837--849},
  year={2024},
  publisher={Nature Publishing Group US New York}
}

@article{kalisch2024autopet,
  title={Autopet III challenge: Incorporating anatomical knowledge into nnUNet for lesion segmentation in PET/CT},
  author={Kalisch, Hamza and H{\"o}rst, Fabian and Herrmann, Ken and Kleesiek, Jens and Seibold, Constantin},
  journal={arXiv preprint arXiv:2409.12155},
  year={2024}
}

@article{nachar2008mann,
  title={The Mann-Whitney U: A test for assessing whether two independent samples come from the same distribution},
  author={Nachar, Nadim and others},
  journal={Tutorials in quantitative Methods for Psychology},
  volume={4},
  number={1},
  pages={13--20},
  year={2008}
}

@inproceedings{gal2016dropout,
  title={Dropout as a bayesian approximation: Representing model uncertainty in deep learning},
  author={Gal, Yarin and Ghahramani, Zoubin},
  booktitle={international conference on machine learning},
  pages={1050--1059},
  year={2016},
  organization={PMLR}
}

@article{malinin2018predictive,
  title={Predictive uncertainty estimation via prior networks},
  author={Malinin, Andrey and Gales, Mark},
  journal={Advances in neural information processing systems},
  volume={31},
  year={2018}
}

@software{meakin_2025_17701815,
  author       = {Meakin, James and
                  Gerke, Paul K. and
                  Kerkstra, Sjoerd and
                  Koopman, Thomas and
                  Mickan, Anne and
                  van Run, Chris and
                  van Zeeland, Harm and
                  Ciompi, Francesco and
                  Hering, Alessa and
                  Jacobs, Colin and
                  Khalili, Nadieh and
                  Koopmans, Peter and
                  van der Laak, Jeroen and
                  Litjens, Geert and
                  Quax, Silvan and
                  Sánchez, Clara I. and
                  Tannhauser, Jos and
                  Groeneveld, Miriam and
                  Huisman, Henkjan},
  title        = {Grand-Challenge.org},
  month        = nov,
  year         = 2025,
  publisher    = {Zenodo},
  version      = {v2025.10},
  doi          = {10.5281/zenodo.17701815},
  url          = {https://doi.org/10.5281/zenodo.17701815},
  swhid        = {swh:1:dir:135994ce78330975372798ac1edccf5f6f55b82b
                   ;origin=https://doi.org/10.5281/zenodo.3356819;vis
                   it=swh:1:snp:416f92ac0637ca8244c006033b3e66a8f0990
                   464;anchor=swh:1:rel:dcf13e7cb9f26c7f7a62b0c631e6b
                   e9ffabc6b87;path=DIAGNijmegen-grand-
                   challenge-204360d
                  },
}

@inproceedings{zhao2022efficient,
  title={Efficient Bayesian uncertainty estimation for nnU-Net},
  author={Zhao, Yidong and Yang, Changchun and Schweidtmann, Artur and Tao, Qian},
  booktitle={International Conference on Medical Image Computing and Computer-Assisted Intervention},
  pages={535--544},
  year={2022},
  organization={Springer}
}

@article{schott2025uncertainty,
  title={Uncertainty quantification for deep learning-based metastatic lesion segmentation on whole body pet/ct},
  author={Schott, Brayden and Santoro-Fernandes, Victor and Klane{\v{c}}ek, {\v{Z}}an and Perlman, Scott and Jeraj, Robert},
  journal={Physics in Medicine \& Biology},
  volume={70},
  number={11},
  pages={115009},
  year={2025},
  publisher={IOP Publishing}
}

@inproceedings{depeweg2018decomposition,
  title={Decomposition of uncertainty in Bayesian deep learning for efficient and risk-sensitive learning},
  author={Depeweg, Stefan and Hernandez-Lobato, Jose-Miguel and Doshi-Velez, Finale and Udluft, Steffen},
  booktitle={International conference on machine learning},
  pages={1184--1193},
  year={2018},
  organization={PMLR}
}

@article{gal2016uncertainty,
  title={Uncertainty in deep learning},
  author={Gal, Yarin and others},
  year={2016},
  publisher={phd thesis, University of Cambridge}
}

@article{blau1962fluorine,
  title={Fluorine-18: a new isotope for bone scanning},
  author={Blau, MONTE and Nagler, WILLIBALD and Bender, MA},
  journal={J. nuclear Med.},
  volume={3},
  year={1962},
  publisher={Roswell Park Memorial Inst., Buffalo}
}

@article{abdar2021review,
  title={A review of uncertainty quantification in deep learning: Techniques, applications and challenges},
  author={Abdar, Moloud and Pourpanah, Farhad and Hussain, Sadiq and Rezazadegan, Dana and Liu, Li and Ghavamzadeh, Mohammad and Fieguth, Paul and Cao, Xiaochun and Khosravi, Abbas and Acharya, U Rajendra and others},
  journal={Information fusion},
  volume={76},
  pages={243--297},
  year={2021},
  publisher={Elsevier}
}

@article{ruxton2006unequal,
  title={The unequal variance t-test is an underused alternative to Student's t-test and the Mann--Whitney U test},
  author={Ruxton, Graeme D},
  journal={Behavioral Ecology},
  volume={17},
  number={4},
  pages={688--690},
  year={2006},
  publisher={Oxford University Press}
}

@article{keskar2016large,
  title={On large-batch training for deep learning: Generalization gap and sharp minima},
  author={Keskar, Nitish Shirish and Mudigere, Dheevatsa and Nocedal, Jorge and Smelyanskiy, Mikhail and Tang, Ping Tak Peter},
  journal={arXiv preprint arXiv:1609.04836},
  year={2016}
}

@inproceedings{glorot2010understanding,
  title={Understanding the difficulty of training deep feedforward neural networks},
  author={Glorot, Xavier and Bengio, Yoshua},
  booktitle={Proceedings of the thirteenth international conference on artificial intelligence and statistics},
  pages={249--256},
  year={2010},
  organization={JMLR Workshop and Conference Proceedings}
}

@article{wang2024dual,
  title={Dual channel CW nnU-Net for 3D PET-CT Lesion Segmentation in 2024 autoPET III Challenge},
  author={Wang, Ching-Wei and Su, Ting-Sheng and Liu, Keng-Wei},
  journal={arXiv preprint arXiv:2409.07144},
  year={2024}
}

@article{chan2024autopet,
  title={AutoPET Challenge: Tumour Synthesis for Data Augmentation},
  author={Chan, Lap Yan Lennon and Li, Chenxin and Yuan, Yixuan},
  journal={arXiv preprint arXiv:2409.08068},
  year={2024}
}

@misc{Ingrisch2024_autoPETIII,
  author       = {Ingrisch, Michael and Dexl, Jakob and Jeblick, Katharina and Cyran, Clemens and Gatidis, Sergios and Kuestner, Thomas},
  title        = {Automated Lesion Segmentation in Whole-Body PET/CT - Multitracer Multicenter Generalization},
  year         = {2024},
  howpublished = {Zenodo, Version 1},
  doi          = {10.5281/zenodo.10990932},
  note         = {Available at \url{https://zenodo.org/records/10990932}},
}

@article{gatidis2024results,
  title={Results from the autoPET challenge on fully automated lesion segmentation in oncologic PET/CT imaging},
  author={Gatidis, Sergios and Fr{\"u}h, Marcel and Fabritius, Matthias P and Gu, Sijing and Nikolaou, Konstantin and Foug{\`e}re, Christian La and Ye, Jin and He, Junjun and Peng, Yige and Bi, Lei and others},
  journal={Nature Machine Intelligence},
  volume={6},
  number={11},
  pages={1396--1405},
  year={2024},
  publisher={Nature Publishing Group UK London}
}

@inproceedings{ronneberger2015u,
  title={U-net: Convolutional networks for biomedical image segmentation},
  author={Ronneberger, Olaf and Fischer, Philipp and Brox, Thomas},
  booktitle={International Conference on Medical image computing and computer-assisted intervention},
  pages={234--241},
  year={2015},
  organization={Springer}
}

@inproceedings{hartmann2021bayesian,
  title={Bayesian u-net for segmenting glaciers in sar imagery},
  author={Hartmann, Andreas and Davari, Amirabbas and Seehaus, Thorsten and Braun, Matthias and Maier, Andreas and Christlein, Vincent},
  booktitle={2021 IEEE International Geoscience and Remote Sensing Symposium IGARSS},
  pages={3479--3482},
  year={2021},
  organization={IEEE}
}

@article{sensoy2018evidential,
  title={Evidential deep learning to quantify classification uncertainty},
  author={Sensoy, Murat and Kaplan, Lance and Kandemir, Melih},
  journal={Advances in neural information processing systems},
  volume={31},
  year={2018}
}

@article{he2023whole,
  title={Whole-body tumor segmentation from PET/CT images using a two-stage cascaded neural network with camouflaged object detection mechanisms},
  author={He, Jiangping and Zhang, Yangjie and Chung, Maggie and Wang, Michael and Wang, Kun and Ma, Yan and Ding, Xiaoyang and Li, Qiang and Pu, Yonglin},
  journal={Medical Physics},
  volume={50},
  number={10},
  pages={6151--6162},
  year={2023},
  publisher={Wiley Online Library}
}

@article{xu2023automatic,
  title={Automatic segmentation of prostate cancer metastases in PSMA PET/CT images using deep neural networks with weighted batch-wise dice loss},
  author={Xu, Yixi and Klyuzhin, Ivan and Harsini, Sara and Ortiz, Anthony and Zhang, Shun and B{\'e}nard, Fran{\c{c}}ois and Dodhia, Rahul and Uribe, Carlos F and Rahmim, Arman and Ferres, Juan Lavista},
  journal={Computers in Biology and Medicine},
  volume={158},
  pages={106882},
  year={2023},
  publisher={Elsevier}
}

@article{li2024automated,
  title={An automated deep learning-based framework for uptake segmentation and classification on PSMA PET/CT imaging of patients with prostate cancer},
  author={Li, Yang and Imami, Maliha R and Zhao, Linmei and Amindarolzarbi, Alireza and Mena, Esther and Leal, Jeffrey and Chen, Junyu and Gafita, Andrei and Voter, Andrew F and Li, Xin and others},
  journal={Journal of Imaging Informatics in Medicine},
  volume={37},
  number={5},
  pages={2206--2215},
  year={2024},
  publisher={Springer}
}

@article{hofman2016we,
  title={How we read oncologic FDG PET/CT},
  author={Hofman, Michael S and Hicks, Rodney J},
  journal={Cancer Imaging},
  volume={16},
  number={1},
  pages={35},
  year={2016},
  publisher={Springer}
}

@article{som1980fluorinated,
  title={A fluorinated glucose analog, 2-fluoro-2-deoxy-D-glucose (F-18): nontoxic tracer for rapid tumor detection},
  author={Som, P and Atkins, HL and Bandoypadhyay, Do and Fowler, JS and MacGregor, RR and Matsui, K and Oster, ZH and Sacker, DF and Shiue, CY and Turner, H and others},
  journal={Journal of Nuclear Medicine},
  volume={21},
  number={7},
  pages={670--675},
  year={1980},
  publisher={Society of Nuclear Medicine}
}

@article{silver1997prostate,
  title={Prostate-specific membrane antigen expression in normal and malignant human tissues.},
  author={Silver, David A and Pellicer, Inmaculada and Fair, William R and Heston, WD and Cordon-Cardo, Carlos},
  journal={Clinical cancer research: an official journal of the American Association for Cancer Research},
  volume={3},
  number={1},
  pages={81--85},
  year={1997}
}

@article{lakshminarayanan2017simple,
  title={Simple and scalable predictive uncertainty estimation using deep ensembles},
  author={Lakshminarayanan, Balaji and Pritzel, Alexander and Blundell, Charles},
  journal={Advances in neural information processing systems},
  volume={30},
  year={2017}
}

@article{kendall2017uncertainties,
  title={What uncertainties do we need in bayesian deep learning for computer vision?},
  author={Kendall, Alex and Gal, Yarin},
  journal={Advances in neural information processing systems},
  volume={30},
  year={2017}
}

@article{kovacs2024data,
  title={Data-Centric Strategies for Overcoming PET/CT Heterogeneity: Insights from the AutoPET III Lesion Segmentation Challenge},
  author={Kovacs, Balint and Xiao, Shuhan and Rokuss, Maximilian and Ulrich, Constantin and Isensee, Fabian and Maier-Hein, Klaus H},
  journal={arXiv preprint arXiv:2409.10120},
  year={2024}
}

@article{der2009aleatory,
  title={Aleatory or epistemic? Does it matter?},
  author={Der Kiureghian, Armen and Ditlevsen, Ove},
  journal={Structural safety},
  volume={31},
  number={2},
  pages={105--112},
  year={2009},
  publisher={Elsevier}
}

@article{roberts2025prospective,
  title={A prospective, multi-centre trial of PSMA-PET compared to FDG-PET for staging of newly diagnosed high risk prostate cancer},
  author={Roberts, Matthew J and Roberts, Natasha A and Pelecanos, Anita and Yaxley, John W and Harley, Simon JD and Siriwardana, Amila R and Cullen, Karla and Prior, Marita and Lindsay, Karen and Vela, Ian and others},
  journal={EJNMMI research},
  volume={15},
  number={1},
  pages={92},
  year={2025},
  publisher={Springer}
}

@article{trotter2023positron,
  title={Positron emission tomography (PET)/computed tomography (CT) imaging in radiation therapy treatment planning: a review of PET imaging tracers and methods to incorporate PET/CT},
  author={Trotter, Jacob and Pantel, Austin R and Teo, Boon-Keng Kevin and Escorcia, Freddy E and Li, Taoran and Pryma, Daniel A and Taunk, Neil K},
  journal={Advances in radiation oncology},
  volume={8},
  number={5},
  pages={101212},
  year={2023},
  publisher={Elsevier}
}





% \end{thebibliography}

\end{document}